\PassOptionsToPackage{table}{xcolor}
\documentclass[]{paper}
\usepackage{ragged2e} 
\RequirePackage{tocloft}
\usepackage{wrapfig}
\usepackage{adjustbox}
\usepackage[utf8]{inputenc}
\usepackage[T1]{fontenc}
\usepackage{hyperref}
\usepackage{url}
\usepackage{booktabs}
\usepackage{amsfonts}
\usepackage{nicefrac}
\usepackage{microtype}
\usepackage{caption}
\usepackage{multirow}
\usepackage{makecell}
\usepackage{siunitx}
\usepackage{wrapfig}
\newsavebox{\tablebox}
\newlength{\tableht}

\usepackage[table]{xcolor}   
\usepackage{booktabs}
\usepackage{array}

\definecolor{stagebg}{HTML}{F2F4F6}   %
\definecolor{okbg}{HTML}{E4F4EA}      %
\definecolor{badbg}{HTML}{FBE8E6}     %
\definecolor{rulegray}{HTML}{C9CFD4}  %
\definecolor{cLogos}{HTML}{1F6FB2}    %
\definecolor{cEvid}{HTML}{1E8A57}     %
\definecolor{cEthos}{HTML}{8E44AD}    %
\definecolor{cPathos}{HTML}{C0392B}   %

\usepackage{graphicx}
\usepackage{amsmath}

\usepackage{fontawesome5}
\newcommand\project[1]{\metadata[GitHub \faGithub]{#1}}

\usepackage{times}
\usepackage[most]{tcolorbox}
\usepackage{latexsym}
\usepackage{microtype}
\usepackage{inconsolata}
\usepackage{graphicx}
\usepackage{amsmath}
\usepackage{amssymb}
\usepackage{booktabs}
\usepackage{multirow}
\usepackage{algorithm}
\usepackage{algpseudocode}
\usepackage{xcolor}
\usepackage{enumitem}
\usepackage{tikz}
\usetikzlibrary{arrows.meta, positioning, shapes.geometric, fit}
\usepackage[utf8]{inputenc}

\usepackage{arydshln}
\usepackage{amssymb}  
\usepackage{pifont}   

\usepackage{xspace}
\usepackage{tabularx}
\usepackage{booktabs}
\usepackage{makecell}

\usepackage{colortbl}
\colorlet{snipbg}{gray!7}
\colorlet{okbg}{green!8}
\colorlet{badbg}{red!8}
\colorlet{samebg}{orange!10}

\usepackage{subcaption}
\usepackage{wrapfig}
\usepackage{listings}
\usepackage{csquotes}

\usepackage{tikz}
\usepackage[edges]{forest}

\title{\textsc{Argus}: Theory-of-Mind Guided Argument Generation with Strategy-Aware Planning and Knowledge Grounding}

\author{\textbf{Zhe Hu}}

\affiliation[1]{InspireOmni AI}
\affiliation[2]{The Hong Kong Polytechnic University} 

\abstract{

Persuasive argument generation requires modeling audience beliefs, rhetorical strategies, and factual grounding. Despite recent advancements, existing methods remain largely audience-agnostic and fail to integrate strategy selection to improve persuasiveness. To bridge this gap, we propose \textsc{Argus}, an agent-based framework that operationalizes classical rhetoric for persuasive writing. At its core, a Theory-of-Mind (ToM) Reasoner constructs an explicit dual mental model of the audience’s beliefs and values to guide downstream decisions. This representation conditions a component-aware planner that decomposes the argument into subtopics, assigns fine-grained rhetorical functions (logos, pathos, ethos), and triggers strategy-guided evidence retrieval at planning time. Finally, a refinement module iteratively targets and resolves multi-dimensional weaknesses without quality regression. We evaluate \textsc{Argus} across three diverse benchmarks using both automated pairwise Elo and LLM-as-judge metrics. Results show that \textsc{Argus} consistently outperforms strong baselines across multiple backbone models, achieving top rankings and highest overall scores. Targeted simulation experiments further validate its effectiveness in shifting resistant audience stances.
}

\Contact{\href{zhehu.derek@gmail.com}{zhehu.derek@gmail.com}}

\project{\href{https://github.com/Derekkk/Argus_Arggen}{https://github.com/Derekkk/Argus\_Arggen}}

\begin{document}

\maketitle


\section{Introduction}

The capacity to generate persuasive arguments underlies a wide range of AI applications, including writing assistance~\cite{ding-etal-2025-feat,bao-etal-2022-aeg}, policy analysis~\cite{verma-etal-2026-predicting}, conversational tools~\cite{wang-etal-2019-persuasion,priya-etal-2025-argue}, and competitive debating systems~\cite{slonim2021autonomous}. As Large Language Models (LLMs) are increasingly deployed in communication-intensive roles~\cite{dyachenko2025llm}, argumentation has transitioned from a downstream utility to a foundational capability. True persuasion, however, is not merely an exercise in surface fluency. It demands a sophisticated convergence of multi-layered cognitive processes: understanding the baseline psychological profile of the audience, selecting calibrated rhetorical strategies that resonate with their systemic values, and anchoring these structures in precise, verifiable evidence~\cite{deane2015key,falk2023storyarg}.  

Early neural approaches collapsed this intricate cognitive pipeline into end-to-end sequential generation, frequently yielding arguments that are structurally incoherent or rhetorically shallow~\cite{hua-etal-2019-argument-generation,wang2023argument}. To address these structural limitations, recent paradigms have pivoted toward agentic workflows leveraging the power of LLMs~\cite{hinton2023persuasive}. These frameworks decompose long-form writing into content planning and writing phases, or leverage multi-agent, debate-driven self-play to iteratively polish contents~\cite{hu-etal-2025-debate,han2026drpg,zhao-etal-2025-plan,hu-etal-2024-americano-argument,xiao-etal-2024-prove}, which improves structural coherence over single-step generation.

While these methods markedly improve surface-level fluency and logical consistency, they have several limitations that impede genuine persuasive efficacy: (1) \textbf{Audience Agnosticism}: Persuasion is inherently relational, yet existing methods lack the computational architecture to explicitly model the dynamic mental states, deep-seated emotional resistance, and value alignment of the receiver, limiting the persuasiveness; (2) \textbf{Strategy-Agnostic Content Planning}: Current planners focus predominantly on topical ordering (i.e., deciding what to say next) while largely ignoring how to say it. They fail to operationalize classical rhetorical modes (e.g., logos, pathos, ethos) as explicit, structurally balanced design objects~\cite{sichach2024ethos,higgins2012ethos,wang-etal-2017-winning}; (3) \textbf{Disjointed Evidence Retrieval}: Fact-grounding in existing frameworks is typically executed either as a coarse-grained, document-level preprocessing step or as a post-hoc patch during drafting~\cite{yeginbergen-etal-2025-dynamic,li2025r}. Consequently, retrieval cannot dynamically shape or adapt to the evolving rhetorical and planning demands of specific subtopics~\cite{gleason1999role}.

In this work, we introduce \textsc{Argus} (\textbf{A}udience-aware \textbf{R}hetorical \textbf{G}eneration with \textbf{U}nified \textbf{S}trategic planning), a novel agent framework that bridges cognitive psychology, classical rhetoric, and agentic language modeling. \textsc{Argus} formalizes argument generation not as an isolated text-production task, but as an explicit, closed-loop structural optimization problem. Our framework introduces a paradigm shift by separating audience profile modeling from generation. Before writing begins, a dedicated Theory-of-Mind (ToM) Reasoner externalizes an explicit, dual mental model capturing the audience's argumentative profiles, emotional triggers, and value landscapes through open-ended natural language phrases. This rich cognitive representation is directly consumed by a Component-Aware Planner. Rather than drafting a flat topical outline, the planner orchestrates the argument at the granular level of rhetorical components, intentionally mapping subtopics to explicit persuasive modes (logos, pathos, ethos) based on estimated audience reactance. Crucially, we integrate strategy-guided evidence retrieval directly into the planning process; web-search queries are generated per subtopic and conditioned on its specific rhetorical goal, embedding empirical grounding into the very blueprint of the argument. Finally, a surgical refiner module is introduced to diagnose targeted flaws and revise final argument.

We conduct extensive evaluations of \textsc{Argus} across three distinct datasets spanning diverse discourse styles: ChangeMyView (CMV), iDebate, and ExplaGraphs. Comprehensive round-robin evaluations using both pairwise Elo ratings and absolute LLM judges confirm that \textsc{Argus} consistently and substantially outperforms competitive baselines across three standard backbone LLM families. To further model persuasive effects, we introduce
a targeted simulation experiments where judges role-play as resistant counterparts, and the results demonstrate that \textsc{Argus} excels at inducing genuine stance shifts in skeptical environments.  In summary, our primary contributions are:
\begin{itemize}[leftmargin=*, topsep=2pt, itemsep=1pt]
\item \textbf{The Theory-of-Mind Computational Mechanism}: We present \textsc{Argus}, an  argument generation framework that integrates explicit Theory-of-Mind reasoning to govern subtopic decomposition, rhetorical alignment, and inline counterargument preemption.
\item \textbf{Component-Aware Rhetorical Planner}: We introduce an advanced planning protocol that maps subtopics to classical rhetorical modes and couples retrieval with localized strategic goals, grounding structure in empirical evidence at planning time.
\item \textbf{Extensive Empirical Validation}: We conduct extensive evaluations across three datasets. Beyond achieving state-of-the-art Elo advantages, we validate through targeted  simulations that explicit ToM profiles translate directly into actual rhetorical influence on resistant human-like audiences.
\end{itemize}

\section{Related Work}
\noindent\textbf{Argument Generation.}
Automatic argument generation has progressed from end-to-end neural models toward increasingly structured, multi-stage pipelines~\cite{hua-wang-2018-neural,wang2023argument,jin-etal-2026-arggenbench}. To improve logical coherence, prior work decomposes generation through explicit text planning and trains models in a multi-task fashion~\cite{hua-etal-2019-argument-generation,hua2021dyploc,schiller2021aspect,bao-etal-2022-aeg}. 
With the advent of LLMs, attention has shifted toward prompting and agent-based methods~\cite{li2025large}. Inspired by chain-of-thought prompting~\cite{wei2022chain}, recent work designs agentic pipelines with decomposition workflows~\cite{hu-etal-2024-americano-argument,xiao-etal-2024-prove,han2026drpg} or multi-agent debate~\cite{hu-etal-2025-debate,zhao-etal-2025-plan} to improve argument quality. 
However, these systems treat the audience as implicit context and plan primarily over content ordering. \textsc{Argus} departs from prior planning-and-debate systems by introducing an explicit audience model for Theory-of-Mind reasoning and by planning at the granularity of rhetorical function rather than content order.

\begin{figure*}[t]
\centering
\includegraphics[scale=0.52]{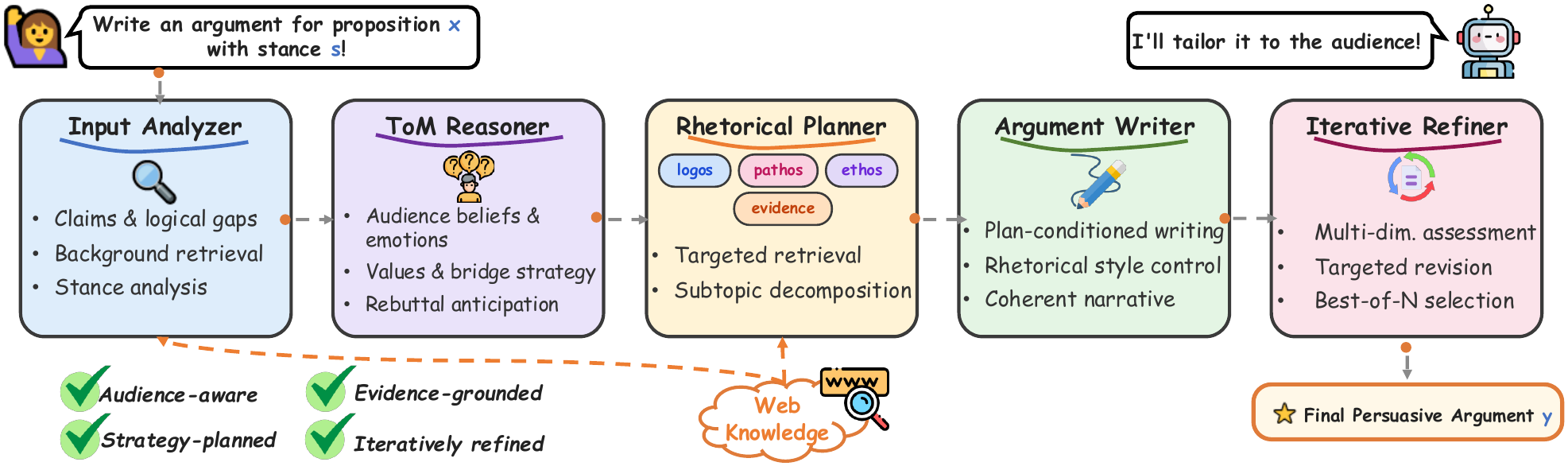}
\caption{Overview of the \textsc{Argus} framework, which consists of five modules for argument generation. }
\label{fig:architecture}
\end{figure*}

\smallskip\noindent\textbf{Theory of Mind and Audience Modeling.}
Argumentation theory holds that persuasion depends jointly on the argument, its source, and the audience~\cite{andriessen1999planning,deane2015key}. Recent work probes whether LLMs possess Theory of Mind (ToM)~\cite{nguyen2025survey}, finding that models exhibit non-trivial but brittle ToM, and that most evaluations test only spectatorial belief prediction rather than the planning ToM needed to deliberately shift a stance~\cite{moore2025large,moore2026large}.
Closer to our setting, a recent line of work models ToM specifically for persuasion: benchmarks for ToM in persuasive dialogue~\cite{fyu2025persuasivetom}, opponent-aware persuaders trained with ToM~\cite{han2025tomap}, dual-knowledge and meta-cognitive multi-agent persuasion frameworks~\cite{ma2026think,zhang2026ma2pmetacognitiveautonomousintelligent}.
In parallel, studies of model persuasiveness show that LLM-generated arguments can rival human-written ones and improve when tailored to the target~\cite{durmus2024persuasion,verma-etal-2026-predicting}. These findings motivate modeling audience mental states  with an explicit ToM module for persuasive argument generation.

\section{Method}

\subsection{Task Formulation}
\label{sec:task}

\label{sec:method}

We study the task of \textit{argument generation}: given a proposition $x$~\footnote{$x$ can be either a short topic claim or a long-form, opinion-rich argumentative passage.} on a controversial topic and a stance $s \in \{\textit{support}, \textit{refute}\}$, write an argumentative article $y$ that persuasively defends $s$ with respect to $x$. This is typically formalized as a conditional text generation problem:
\begin{equation}
y = \mathcal{M}(x, s),
\end{equation}
where $\mathcal{M}$ denotes the generation framework. Rather than treating $\mathcal{M}$ as a single prompting step over an LLM, we decompose it into a sequence of interpretable reasoning and generation actions. This decomposition enables key dimensions of argument generation, including audience alignment, rhetorical organization, evidentiary grounding, and linguistic coherence, to be modeled explicitly and optimized separately.

\subsection{System Overview}
\label{sec:overview}

The \textsc{Argus} framework operationalizes this decomposition through five sequential modules, as illustrated in Figure \ref{fig:architecture}: Input Analyzer, ToM Reasoner, Argument Planner, Argument Writer, and Iterative Refiner. Each module operates on structured representations, and outputs are  passed to the downstream components.

\subsection{Input Analysis}
\label{sec:input}

Effective argument generation begins with understanding the input proposition itself~\cite{seo2023good,deane2015key}. Human debaters and persuasive writers rarely start drafting immediately; instead, they first analyze the logical structure of the input, identify implicit assumptions, and gather relevant background knowledge, especially when the input is a complete argument rather than a short claim. \textsc{Argus} follows this principle through an \textit{Input Analysis} stage that produces structured representations for subsequent stages.

Concretely, the Input Analyzer first decides whether external knowledge is required and, if so, generates targeted web queries whose retrieved results are incorporated into the input context. This retrieval-first design ensures that the subsequent analysis is grounded in actual evidence rather than parametric knowledge alone. It then produces a structured representation: (i) \textit{key claims} stated in the input; (ii) \textit{background knowledge} synthesized from retrieved snippets; and (iii) an optional \textit{logical structure} decomposition, including premises, implicit assumptions, and potential logical gaps, when the input is a developed argument rather than a short proposition.

A concrete example is shown in Figure~\ref{fig:input-analysis-example}. This structured representation guides downstream planning by providing grounded contextual knowledge and by exposing implicit assumptions and logical gaps, enabling the Planner to construct targeted counterarguments or focus on weak points of the opposing position.

\subsection{Theory-of-Mind Reasoning}

A persistent limitation of existing argument generation systems is that the audience is treated as an implicit context coupled in the generation process, never as a structured object that reasoning can operate over. However, persuasive skills require writers to anticipate the attitudes, beliefs, and arguments of the audience in order to fully engage the reader in the argument~\cite{deane2015key,andriessen1999planning}.
\textsc{Argus} addresses this with an explicit Theory-of-Mind (ToM) Reasoner that constructs a dual mental model $\Psi$ of the target audience. By incorporating this module, the subsequent planner can use the audience model to make 
structural decisions, including which subtopics to include, which rhetorical components to assign, and which anticipated objections to rebut inline. 

Formally, the ToM Reasoner takes as input the proposition, stance, and input analysis to produce a dual mental model $\Psi = (\Psi_\mathcal{O},
\Psi_\mathcal{V})$. The \textit{opponent model} $\Psi_\mathcal{O}$ captures
the audience's argumentative profile: the emotional themes that resonate with them, and a set of specific positions they hold, each paired with the underlying belief or motivation driving it. The \textit{value model}
$\Psi_\mathcal{V}$ represents the audience's value landscape as open-ended natural language phrases~\cite{sorensen2024value} (e.g., \textit{``desire
for personal safety''}), along
with bridging strategies that identify how the argument's position can be reconciled with values that may initially conflict with it. An example is shown in Figure~\ref{fig:tom-example}.

\subsection{Argument Planning}
\label{sec:planner}

Given the input analysis and the audience model outputs, the Argument Planner constructs a rhetorical blueprint $\mathcal{P}$ for generation, which is represented as a structured sequence of subtopics $\mathcal{P} = \langle t_1, t_2, \dots, t_n \rangle$. Each subtopic $t_i$ is represented as:
\begin{equation}
\text{Attr}(t_i) = (c_i, k_i, e_i, r_i),
\end{equation}
where $c_i \in \{\textit{logos, pathos, ethos, evidence}\}$ denotes the primary rhetorical component assigned to $t_i$ , $k_i$ is the set of key claims , $e_i$ represents the optionally retrieved evidence grounding the subtopic , and $r_i$ is an anticipated audience rebuttal for inline preemption.

\paragraph{Rhetorical Component Planning.}
Existing text planning approaches primarily focus on topical decomposition and content ordering~\cite{hua-etal-2021-dyploc,he-etal-2024-decomposing}. However,  human writers and debaters naturally exercise precise, \textit{strategic control over rhetorical functions and audience adaptation} to maximize persuasive resonance~\cite{wang-etal-2017-winning,wachsmuth2018argumentation}. To bridge this gap, \textsc{Argus} deliberately plans not only \emph{what} to articulate, but also \emph{how} each discrete subtopic should persuade.

Specifically, subtopics centered on logical reasoning are assigned \textit{logos}; emotionally salient concerns are assigned \textit{pathos}; appeals to credibility, authority, or shared norms are assigned \textit{ethos}; and empirically grounded claims are assigned \textit{evidence}~\cite{jung2026argumentation}. This allocation is adaptively conditioned on the audience model $\Psi$. For instance, audiences projected to exhibit high ideological resistance or acute emotional sensitivity are met with elevated \textit{pathos}- and \textit{ethos}-driven strategies to minimize psychological reactance, whereas analytically inclined audiences are targeted with  structurally rigorous \textit{logos}- and \textit{evidence}-driven frameworks.

\paragraph{Strategy-Guided Evidence Retrieval.}
To anchor persuasive rhetoric in factual reality, evidence retrieval is integrated directly into the planning process to ensure factual grounding. For subtopics that require external grounding, the planner dynamically generates targeted search queries to fetch context-specific knowledge. This subtopic-level execution contrasts sharply with conventional proposition-level retrieval, which frequently scatters an undifferentiated pool of background documents uniformly across an entire text. By tying evidence directly to the local rhetorical function $c_i$ and local claims $k_i$, our method explicitly enhances the verifiability of individual sub-topics. This fine-grained verification serves a critical fact-checking role during planning, ensuring that the generated narrative is authoritative, reliable, and fundamentally persuasive~\cite{chen-etal-2024-complex}.

\paragraph{Iterative Plan Self-Evaluation.} 
After initial plan generation, the planner evaluates its own output against criteria including rhetorical diversity, ToM alignment, logical ordering, and stance consistency. If the plan does not meet a quality threshold, it is revised and evidence is re-retrieved for any modified subtopics. This process repeats multiple iterations, ensuring the blueprint is both rhetorically sound and empirically grounded before writing begins.

\subsection{Argument Writing and Refinement}
\label{sec:writer}
Given the structured plan $\mathcal{P}$, the Argument Writer produces the full argument. This stage strictly preserves the intended rhetorical composition, local evidence grounding, and audience value framing established in the planning phase. This ensures that the final output is both globally controllable and contextually persuasive.
To maximize persuasive efficacy and eliminate structural flaws, \textsc{Argus} integrates a multi-dimensional iterative refinement loop directly into the text generation pipeline~\cite{madaan2023self,hu-etal-2024-americano-argument}. Once the draft is produced, an LLM-based evaluator assesses the argument along multiple dimensions, including persuasiveness, coherence, factual accuracy, value alignment, and rhetorical balance, yielding an overall quality score. If the score falls below a threshold, the system identifies the weakest dimensions and generates targeted revision instructions focused on those specific deficiencies. 

These \textit{targeted} revisions can effectively address specific deficiencies while preserving working portions of the draft. To guard against quality regression, where improvements in one aspect inadvertently degrade others, the system maintains a best-of-$n$ buffer across refinement rounds, returning the highest-scoring version if a later revision fails to improve overall quality.

\subsection{Design Rationale}
\label{sec:design}

The design decisions warrant justification. First, ToM reasoning is separated from planning rather than collapsed into a single prompt: externalizing $\Psi$ ensures the Planner conditions on a fully elaborated audience mental model, and its key signals, anticipated objections and value alignment notes, are distilled into each subtopic's structured fields, making them available to the Writer without an additional inference step. Second, rhetorical component assignment happens at planning time rather than being delegated to the Writer, making the argument's persuasive structure an explicit, inspectable object that can be deliberately balanced across subtopics.

\section{Experimental Setup}
\label{sec:exp}

\subsection{Datasets and Tasks}
\label{sec:setup}

We evaluate \textsc{Argus} on argument generation across three benchmarks that collectively span diverse domains, stances, and discourse styles:
(1) \textbf{ChangeMyView (CMV)}~\citep{hua-etal-2019-argument-generation} consists of Reddit posts on politics and policy domain where the original post (OP) explicitly invites counterarguments. The target stance is \emph{refute}, and we concatenate the title and OP body as input;
(2) \textbf{iDebate}~\citep{hu-etal-2025-debate} comprises short propositions on controversial topics drawn from a broad range of domains. The target stance is \emph{refute}; propositions are typically abstract, making the task substantially more open-ended;
(3) \textbf{ExplaGraphs}~\citep{saha-etal-2021-explagraphs} requires the model to generate an argument that \emph{supports} a given statement. High-quality outputs depend on relevant background knowledge of the debate topic.

\begin{table*}[t]
\fontsize{10}{12}\selectfont
\centering\scriptsize
\begin{tabular}{ll ccccc ccccc ccccc}
\toprule
\multirow{2}{*}{\textbf{Backbone}} & \multirow{2}{*}{\textbf{Method}}
  & \multicolumn{5}{c}{\textbf{CMV}}
  & \multicolumn{5}{c}{\textbf{ExplaGraphs}}
  & \multicolumn{5}{c}{\textbf{iDebate}} \\
\cmidrule(lr){3-7}\cmidrule(lr){8-12}\cmidrule(lr){13-17}
 & & ELO & WR\% & Per. & Coh. & Facc.
   & ELO & WR\% & Per. & Coh. & Facc.
   & ELO & WR\% & Per. & Coh. & Facc. \\
\midrule
\multirow{5}{*}{DeepSeek-v3.2}
  & \textbf{ARGUS}
    & \textbf{1661} & \textbf{67.2} & \textbf{3.94} & \textbf{4.56} & \textbf{3.65}
    & \textbf{1705} & \textbf{62.5} & \textbf{4.17} & \textbf{4.65} & \textbf{3.83}
    & \textbf{1865} & \textbf{81.7} & \textbf{4.13} & \textbf{4.62} & \textbf{3.60} \\
  & Plan\&Write
    & 1458 & 25.0 & 3.28 & 4.48 & 3.39
    & 1592 & 38.3 & 3.97 & 4.60 & 3.48
    & 1478 & 28.3 & 4.00 & 4.60 & 3.46 \\
  & Self-Refine
    & 1507 & 18.1 & 3.50 & 4.48 & 3.50
    & 1539 & 19.2 & 4.03 & 4.58 & 3.54
    & 1365 &  9.2 & 4.00 & 4.58 & 3.48 \\
  & Debate
    & 1479 & 16.4 & 3.29 & 4.43 & 3.38
    & 1214 &  4.2 & 3.68 & 4.47 & 3.08
    & 1422 &  8.3 & 3.88 & 4.54 & 3.27 \\
  & Direct
    & 1396 & 10.3 & 3.16 & 4.41 & 3.38
    & 1451 & 10.8 & 3.94 & 4.61 & 3.38
    & 1370 &  7.5 & 4.04 & 4.61 & 3.43 \\
\midrule
\multirow{5}{*}{Qwen3.5-Flash}
  & \textbf{ARGUS}
    & \textbf{1763} & \textbf{81.0} & \textbf{3.88} & \textbf{4.49} & \textbf{3.49}
    & \textbf{1791} & \textbf{75.8} & \textbf{4.06} & 4.52 & \textbf{3.47}
    & \textbf{1880} & \textbf{89.2} & \textbf{4.01} & \textbf{4.55} & \textbf{3.37} \\
  & Plan\&Write
    & 1457 & 30.2 & 3.18 & 4.33 & 2.97
    & 1567 & 35.0 & 3.95 & 4.50 & 3.03
    & 1480 & 30.8 & 3.93 & 4.53 & 3.18 \\
  & Self-Refine
    & 1516 & 31.9 & 3.07 & 4.30 & 3.21
    & 1448 & 33.3 & 3.85 & 4.41 & 3.22
    & 1495 & 25.8 & 3.36 & 4.20 & 3.05 \\
  & Debate
    & 1441 & 19.8 & 2.28 & 4.28 & 2.97
    & 1231 &  3.3 & 3.42 & 4.21 & 2.88
    & 1228 &  4.2 & 2.87 & 4.30 & 2.93 \\
  & Direct
    & 1323 &  9.5 & 3.52 & 4.34 & 2.99
    & 1464 & 22.5 & 3.96 & \textbf{4.54} & 3.26
    & 1417 & 22.5 & 3.89 & 4.51 & 3.11 \\
\midrule
\multirow{5}{*}{GPT-5-mini}
  & \textbf{ARGUS}
    & \textbf{1699} & \textbf{58.3} & \textbf{4.11} & \textbf{4.64} & \textbf{3.99}
    & \textbf{1704} & \textbf{58.3} & 4.26 & \textbf{4.68} & 4.03
    & \textbf{1804} & \textbf{82.5} & 4.19 & \textbf{4.67} & \textbf{3.90} \\
  & Plan\&Write
    & 1472 & 17.5 & 3.82 & 4.56 & 3.83
    & 1498 & 17.5 & \textbf{4.29} & \textbf{4.68} & 3.95
    & 1509 & 22.5 & \textbf{4.21} & 4.66 & 3.83 \\
  & Self-Refine
    & 1483 & 15.0 & 3.17 & 4.44 & 3.94
    & 1513 & 10.0 & 3.71 & 4.47 & \textbf{4.08}
    & 1491 & 18.3 & 3.57 & 4.49 & 3.87 \\
  & Debate
    & 1557 & 25.8 & 3.54 & 4.56 & 3.89
    & 1501 & 16.7 & 4.21 & 4.65 & 3.87
    & 1491 & 18.3 & 4.14 & 4.64 & 3.70 \\
  & Direct
    & 1290 &  4.2 & 3.80 & 4.55 & 3.68
    & 1285 &  1.7 & 4.17 & \textbf{4.68} & 3.82
    & 1205 &  2.5 & 4.09 & 4.64 & 3.63 \\
\bottomrule
\end{tabular}
\vspace{-2mm}
\caption{Main Results. We report pairwise ELO, win rate (WR\%), and absolute LLM-as-judge scores on persuasiveness (Per.), coherence (Coh.), and factual accuracy (Facc.) on a scale of 0--5. \textbf{Bold} marks the best score.}
\vspace{-4mm}
\label{tab:main}
\end{table*}

\subsection{Baselines}

We compare \textsc{Argus} against four strong baselines with the same backbone LLMs as ours:
\begin{itemize}[leftmargin=*, itemsep=0pt, topsep=0pt, parsep=0pt]
    \item \textbf{Direct}: single-pass generation with a persuasion-focused system instruction.
    \item \textbf{Plan-and-Write}: the model first drafts an argument plan and then realizes it as a full argument.
    \item \textbf{Self-Refine}~\citep{madaan2023self}: the model produces an initial argument and then iteratively critiques and revises its own output.
    \item \textbf{Debate}~\citep{hu-etal-2025-debate}: a multi-agent setup where agents adopt opposing stances to debate and produce the argument. We adopt a simplified two-agent configuration (one agent per stance) followed by a single synthesis pass.
\end{itemize}

\noindent To assess generalizability, we include  three backbone models: DeepSeek-V3.2, Qwen3.5-Flash, and GPT-5-mini. More details are in Appendix~\ref{app:details}.

\subsection{Evaluation Metrics}

We adopt a two evaluation protocol combining pairwise comparison and absolute scoring:

\noindent\textbf{Pairwise Elo evaluation.} Following \citet{elo1967proposed} and \citet{bai2022training}, we conduct round-robin pairwise comparisons among all systems for every input. Each pair is judged twice with the order swapped to mitigate position bias. To prevent noise-driven rating drift, a win is declared only when the judge's score difference exceeds $0.5$; smaller differences are recorded as ties. Elo ratings are updated with $K=32$ and an initial rating of $1500$.

\smallskip
\noindent\textbf{Absolute scoring.} In addition, an LLM judge scores each argument independently along aspects including \textit{persuasiveness}, \textit{coherence}, and \textit{factual accuracy}.

\smallskip\noindent In all settings, the judge is fixed to {GPT-5.4} which is different from the generation backbone~\footnote{We deliberately select GPT-5.4, a model stronger than all three backbones, as the judge, so that evaluation quality is not bottlenecked by the evaluator's own capability.}. The details are in Appendix~\ref{app:eval_details}.

\section{Results and Analysis}
\label{sec:results}

\subsection{Main Results}
Table~\ref{tab:main} reports pairwise ELO rankings and absolute LLM-as-judge scores across all datasets. Overall, \textsc{Argus} consistently outperforms all baselines across evaluation settings.

\smallskip\noindent\textbf{Pairwise Evaluation.}
For pairwise ELO ranking, \textsc{Argus} achieves the most significant gains on iDebate, where inputs are short, open-domain propositions requiring broad world knowledge, open-ended reasoning, and flexible rhetorical construction. This suggests that structured planning and deliberate argument development are particularly beneficial when the input provides limited inherent argumentative scaffolding. Among baselines, Plan\&Write is consistently the strongest competitor, indicating that explicit reasoning can provide a strong improvement over direct generation, while Self-Refine and Multi-Agent Debate lag behind due to instability and insufficient coordination.

\textsc{Argus} also demonstrates strong improvements on CMV, where inputs consist of full, opinion-rich statements. The results suggest that \textsc{Argus} can effectively model discourse structure and leverage ToM-guided planning to tailor arguments to detailed and audience-specific beliefs and preferences. In contrast, baseline methods show noticeable instability across backbones, highlighting the difficulty of relying solely on iterative critic or agent-level deliberation without explicit audience modeling.

Finally, on ExplaGraphs, \textsc{Argus} consistently achieves the best ELO scores across all settings while maintaining strong factual consistency and coherent reasoning. Notably, Self-Refine occasionally improves factuality but remains inconsistent in overall ranking performance. These results demonstrate that the proposed planning framework is effective for both subjective persuasion tasks and more structured, fact-oriented reasoning scenarios.

\smallskip\noindent\textbf{Absolute Evaluation.}
The absolute LLM-as-judge scores largely corroborate the pairwise results. \textsc{Argus} achieves the highest scores across nearly all evaluation dimensions, particularly in persuasiveness and factual accuracy, where it shows clear and consistent margins over baselines. In contrast, coherence scores are uniformly high across methods, suggesting that modern LLMs already produce structurally fluent arguments and that coherence is not a primary differentiating factor.

Interestingly, baseline behavior varies across models and datasets. On ExplaGraphs under GPT-5-mini, Plan\&Write achieves the highest persuasiveness score, while Self-Refine attains the best factual accuracy. However, \textsc{Argus} still wins in pairwise comparisons, indicating that isolated rubric dimensions do not fully capture holistic quality. In particular, arguments with slightly lower individual scores may still be preferred due to better global balance, organization, or rhetorical effectiveness.

Overall, these results highlight that performance differences are primarily driven by persuasiveness and factual grounding, rather than surface-level fluency. The consistent improvements of \textsc{Argus} across datasets and backbones underscore the importance of integrating audience modeling, evidence-grounded planning, and targeted refinement within a unified generation framework.





\subsection{Persuasion Analysis via Target Simulation}


\begin{figure}[t]
\centering
\sbox{\tablebox}{%
  \fontsize{8}{8}\selectfont
  \setlength{\tabcolsep}{1.3mm}
  \begin{tabular}{ll ccc c}
  \toprule
  \textbf{Backbone} & \textbf{Method}
    & \textbf{CMV} & \textbf{ExplaGraphs} & \textbf{iDebate} & \textbf{Avg.} \\
  \midrule
  \multirow{5}{*}{DeepSeek}
    & \textbf{ARGUS} & \textbf{6.71} & \textbf{8.85} & \textbf{8.07} & \textbf{7.88} \\
    &  Plan\&Write            & 5.34 & 8.28 & 7.90 & 7.17 \\
    & Self-Refine    & 6.31 & 8.35 & 7.72 & 7.46 \\
    & Debate         & 5.41 & 8.10 & 7.07 & 6.86 \\
    & Direct         & 5.90 & 8.35 & 7.98 & 7.41 \\
  \midrule
  \multirow{5}{*}{Qwen}
    & \textbf{ARGUS} & \textbf{6.02} & \textbf{7.07} & \textbf{6.47} & \textbf{6.52} \\
    &  Plan\&Write            & 5.12 & 6.38 & 5.90 & 5.80 \\
    & Self-Refine    & 5.03 & 6.48 & 5.58 & 5.70 \\
    & Debate         & 3.45 & 6.00 & 3.90 & 4.45 \\
    & Direct         & 5.31 & 6.22 & 5.70 & 5.74 \\
  \midrule
  \multirow{5}{*}{GPT-5-mini}
    & \textbf{ARGUS} & \textbf{7.45} & \textbf{7.50} & \textbf{8.29} & \textbf{7.75} \\
    &  Plan\&Write            & 7.25 & 7.13 & 7.95 & 7.44 \\
    & Self-Refine    & 6.98 & 7.17 & 7.62 & 7.26 \\
    & Debate         & 7.20 & 7.23 & 8.22 & 7.55 \\
    & Direct         & 6.98 & 7.07 & 8.00 & 7.35 \\
  \bottomrule
  \end{tabular}%
}
\settoheight{\tableht}{\usebox{\tablebox}}  

\begin{minipage}[t]{0.48\textwidth}
  \vspace{0mm}
  \centering
  \usebox{\tablebox}
  \captionof{table}{Persuasion scores (1--10), where higher scores
  indicate more persuasive. Best scores are in \textbf{bold}.}
  \label{tab:persuasion}
\end{minipage}
\hfill
\begin{minipage}[t]{0.48\textwidth}
  \vspace{0mm}
  \centering
  \includegraphics[scale=0.6]{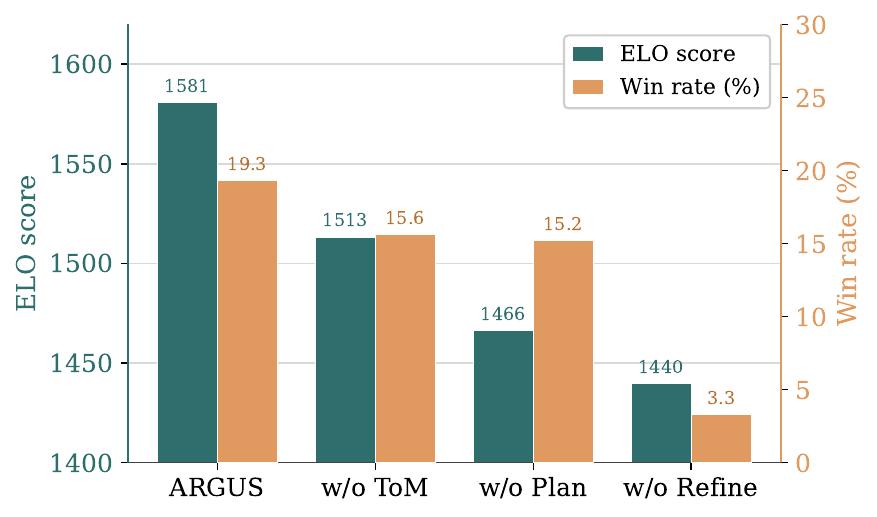}
  \captionof{figure}{Ablation study results (GPT-5-mini backbone,
  ELO pairwise evaluation).}
  \label{fig:ablation}
\end{minipage}
\end{figure}

While pairwise ELO and absolute scores evaluate argument quality from a neutral perspective, the ultimate goal of persuasion is to shift the stance of an opposing audience. To better capture this effect, we introduce a targeted simulation setting where an LLM judge role-plays as a skeptical audience holding the original position.

Specifically, the judge rates each argument on a 1–10 scale, where 1 indicates no change in stance and 10 indicates full persuasion. Each argument is scored twice by independent passes of the judge model, and the final score is the average of the two passes to reduce stochasticity. This setup provides a stricter measure of persuasive impact that is more aligned with real-world persuasion.


As shown in Table~\ref{tab:persuasion}, \textsc{Argus} consistently achieves the highest persuasion scores, confirming that its advantages in pairwise evaluation translate into stronger ability to potentially shift audience stance. The improvement is particularly pronounced on CMV, where inputs contain rich opinions, emotions, and implicit beliefs. This suggests that \textit{\textsc{Argus} effectively exploits such contextual signals to generate more targeted and convincing arguments}.

We also observe that Multi-Agent Debate performs notably worse under the Qwen backbone, indicating that unconstrained multi-agent deliberation may be sensitive to backbone models and can produce coherent but misaligned arguments that fail to address audience-specific concerns, resulting in weaker persuasive impact.

\subsection{Human Evaluation}
\label{sec:human_eval}

\begin{wrapfigure}{tr}{0.45\textwidth}
    \vspace{-4mm}
    \centering
    \includegraphics[scale=0.6]{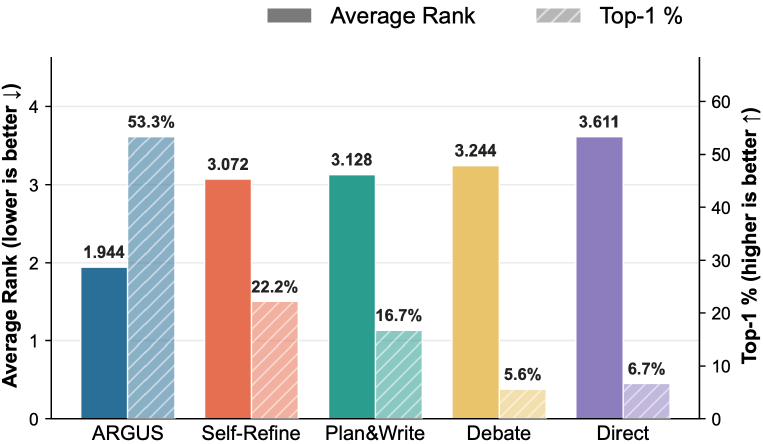}
    \vspace{-3mm}
    \caption{Human evaluation of overall persuasive preference. Overall is the mean rank across
annotators (lower is better); 1st \% is the first-place rate (ties averaged, so totals may slightly exceed 100\%).
    }
    \label{fig:human_eval}
    \vspace{-5mm}
\end{wrapfigure}

To complement our automatic evaluation protocols, we conduct a human evaluation to assess the overall persuasiveness of the generated arguments. We randomly sample 30 propositions and collect arguments generated by ARGUS and the four baselines using the same GPT-5-mini backbone. For each proposition, the five arguments are anonymized and randomly shuffled. Three annotators are hired to rank the arguments according to their overall preference (with ties allowed), and to provide a brief free-text rationale. We convert the rankings into numeric scores, where a lower rank indicates stronger preference, and report the mean rank and first-place rate for each method.

As shown in Figure~\ref{fig:human_eval}, ARGUS achieves the best mean rank and the highest first-place rate, consistently outperforming all four baselines. This finding is consistent with our automatic evaluation results, where ARGUS also achieves the strongest performance across both LLM-judge and persuasion-simulation metrics. The consistent advantage observed across human and automatic evaluations suggests that the gains of ARGUS are not an artifact of a particular evaluation protocol or LLM-based judge, but instead reflect a boader improvement in perceived persuasiveness.

We further analyze the free-text rationales provided by the annotators to better understand the factors underlying their preferences. Several interesting patterns emerge. First, arguments generated by Self-Refine are frequently described as logically clear but relatively flat. For example, annotators characterized them as ``correct but rigid,'' noting that these
arguments often provided clear guidance on \emph{how} to act without creating a strong sense of persuasion. Second, Plan\&Write tends to use analogies and more vivid narration, suggesting that narrative richness can contribute substantially to human-perceived persuasiveness.

The rationales also reveal a distinction between human and automatic evaluation. LLM-based judges tend to place greater emphasis on structural completeness, explicit reasoning, and evidence density, whereas human readers may additionally value narrative appeal, naturalness, and the subjective experience of being persuaded. This difference highlights the importance of strategic planning in ARGUS: by explicitly modeling persuasive
strategies, ARGUS can better balance logical reasoning with rhetorical and narrative qualities that matter to human readers. Importantly, ARGUS remains the most preferred method across both human and automatic evaluation protocols, providing complementary evidence for the effectiveness of its strategy-aware planning.

\subsection{Ablation Study}
\label{sec:ablation}

\begin{wraptable}{tl}{0.46\textwidth}
\centering
\fontsize{8}{9.5}\selectfont
\renewcommand{\arraystretch}{1.15}
\setlength{\tabcolsep}{5pt}
\sisetup{
  table-format=+1.3,
  table-space-text-post={$^*$},
  detect-weight=true,
}
\begin{tabular}{l *{3}{S[table-format=2.1]}}
\toprule
& {\textbf{CMV}} & {\textbf{iDebate}} & {\textbf{ExplaGraphs}} \\
\midrule
\rowcolor{gray!12}
\multicolumn{4}{l}{\textit{Component allocation (\% of subtopics)}} \\
\quad Logos     & 44.1 & 35.9 & 36.9 \\
\quad Pathos    & 19.3 & 23.9 & 19.4 \\
\quad Ethos     & 22.7 & 27.0 & 28.4 \\
\quad Evidence  & 13.8 & 12.0 & 15.3 \\
\midrule
\rowcolor{gray!12}
\multicolumn{4}{l}{\textit{Quality progression (0--5)}} \\
\quad Initial draft ($v_1$)        & {3.49} & {3.69} & {3.79} \\
\quad Final score ($v_\text{fin}$) & {3.96} & {4.07} & {4.11} \\
\quad Refinement gain ($\Delta$)   & {\bfseries +0.47} & {+0.38} & {+0.32} \\
\midrule
\rowcolor{gray!12}
\multicolumn{4}{l}{\textit{Correlations with persuasion}} \\
\quad $r(\text{pathos})$   & {$-0.307^*$} & {$+0.078$} & {$-0.174$} \\
\quad $r(\text{ethos})$    & {$+0.211^*$} & {$-0.044$} & {$-0.018$} \\
\quad $r(\text{evidence})$ & {$+0.122$}   & {$+0.139$} & {$+0.171$} \\
\bottomrule
\end{tabular}
\vspace{-1mm}
\caption{\textsc{Argus} pipeline statistics and Pearson correlations
between rhetorical component allocation and persuasion score
($^*$: $p<0.05$).}
\vspace{-4mm}
\label{tab:pipeline_analysis}
\end{wraptable}

We conduct ablation studies to examine the contribution of each core component in \textsc{Argus} using GPT-5-mini backbone. We evaluate three variants: (1) {w/o ToM}, which removes the ToM Reasoner and thus eliminates audience belief and preference modeling; (2) {w/o Plan}, which bypasses structured argument planning and directly feeds the ToM output to the writer; and (3) {w/o Refine}, which disables iterative refinement. We conduct pairwise evaluation among these four variants.

As shown in Figure~\ref{fig:ablation}, removing any single component degrades the performance, confirming their effectiveness. Meanwhile, we can observe that bypassing the Argument Planner ({w/o Plan}) degrades the win rate more severely than omitting the ToM alone, indicating that rich audience profiles are only marginally effective unless explicitly operationalized through structured rhetorical plans and subtopic decomposition. Finally, removing the Iterative Refiner leads to the large performance decrease, demonstrating the importance of our targeted, multi-dimensional revision.

While removing the Iterative Refiner causes the largest performance drop, refinement alone is insufficient for strong persuasion. Effective refinement depends on high-quality drafts produced by ToM-guided planning, whereas poorly organized drafts limit the benefits of post-hoc refinement. This is supported by the weaker performance of the Self-Refine baseline in Table~\ref{tab:main}, indicating that strategic audience-aware planning provides complementary value that refinement alone cannot replace.

\subsection{Pipeline Component Analysis}

To better understand how intermediate pipeline outputs relate to final argument persuasiveness, we analyze their corresponding scores in Table~\ref{tab:pipeline_analysis}.
The planner exhibits distinct rhetorical preferences across datasets. CMV arguments are substantially more logos-oriented (44\%), consistent with the deliberative nature of the forum, where explicit logical reasoning is expected. In contrast, iDebate and ExplaGraphs place greater emphasis on ethos ($\sim$28\%), as their shorter proposition-style inputs make quickly establishing credibility more effective than developing extended logical chains. Across all three benchmarks, the refinement stage consistently improves argument quality, with the largest gains observed on CMV.

A key finding is that the effectiveness of rhetorical strategies is highly audience-dependent. On {CMV}, rhetorical allocation plays a significant role: pathos-heavy arguments are negatively associated with persuasion, whereas stronger use of ethos correlates positively with effectiveness. 
In contrast, for {iDebate} and {ExplaGraphs}, these correlations largely disappear. The brevity of the propositions provides limited room for differentiated rhetorical strategies, making persuasiveness depend more on the overall quality and coherence of the argument than on the specific rhetorical mix.





\begin{table*}[t]
\centering
\setlength{\tabcolsep}{5pt}
\renewcommand{\arraystretch}{1.25}
\small
\begin{tabular}{@{}p{2.2cm}p{14.2cm}@{}}
\toprule
\textbf{Stage} & \textbf{Successful Case\;\textcolor{green!50!black}{(GPT-5-mini)}} \\
\midrule

\textbf{Task} &
  Write an argument to \textbf{support}: \textit{``Autonomous cars have safety algorithms.''} \\[2pt]

\textbf{ToM: Opp.\ beliefs} &
  \textit{```Safety algorithm' is a marketing phrase used to overstate protection.''}\newline
  \textit{``Algorithms cannot replicate human judgment in novel or ethically fraught situations.''}\newline
  \textit{``Insufficient transparency, independent testing, and accountability for safety claims.''} \\[2pt]

\textbf{Arg.\ Plan} &
  \textcolor{blue!70!black}{[logos]} Define ``safety algorithms'' concretely\newline
  \textcolor{teal!70!black}{[evidence]} Real-world implementations \& independent testing\newline
  \textcolor{blue!70!black}{[logos]} Acknowledge failure modes \& mitigations\newline
  \textcolor{brown!70!black}{[ethos]} Transparency, oversight, accountability \\[4pt]

\midrule
\textbf{Final Argument} &
  \colorbox{snipbg}{\parbox{13.8cm}{\small\itshape``A safety algorithm is simply software designed to detect danger and reduce the chance or severity of harm. With that definition in mind, it's accurate and demonstrable that autonomous cars include safety algorithms. If a vehicle is truly autonomous\ldots it must sense the environment, predict what might happen next, and take actions to avoid or mitigate collisions\ldots These are not marketing phrases. They correspond to implemented features you can point to and test: automatic emergency braking that computes collision risk and applies brakes, collision-avoidance maneuvers that alter a planned route to prevent impact, and sensor-fault detection that forces a safe stop when localization or perception fails.''}} \\[6pt]

\midrule
\textbf{Why it works} &
  \cellcolor{okbg}
  The Plan directly addresses the ToM-identified objection (``marketing phrase'') by defining the term \emph{before} making any factual claim. This definition is then grounded in concrete, testable engineering functions, resolving the audience's conceptual disagreement before introducing supporting evidence. As a result, the generated argument follows a rhetorical structure tailored to the audience's beliefs. \\
  \bottomrule
\end{tabular}
\vspace{-2pt}
\caption{A sample output of ARGUS, linking the ToM audience model and the Planner's rhetorical structure to the delivered final argument. The Plan's ordering directly disarms the audience's stated objection before the argument makes any claim.}
\label{tab:case_study}
\end{table*}

\subsection{Case Study}
\label{sec:case_study}

Table~\ref{tab:case_study} presents a representative successful example, illustrating how the ToM Reasoner's audience model guides the Planner's rhetorical strategy and ultimately shapes the delivered argument.

For the task of writing an argument to \textbf{support} the claim ``Autonomous cars have safety algorithms,'' the ToM Reasoner infers the audience's primary concern: rather than disputing autonomous driving itself, the audience views the term \emph{``safety algorithm''} as an unsubstantiated marketing phrase. Guided by this audience model, the Planner does not immediately present evidence. Instead, it first establishes a precise definition of the term, then introduces concrete engineering implementations and real-world evidence, acknowledges limitations, and finally emphasizes transparency and accountability. This ordering directly addresses the audience's conceptual objection before making factual claims.

The generated argument faithfully follows this plan. It begins by defining ``safety algorithm'' as software designed to detect danger and reduce harm, thereby reframing the disputed concept into a concrete and testable engineering notion. It then supports this definition with examples such as automatic emergency braking, collision-avoidance systems, and sensor-fault detection, before concluding with transparency and oversight considerations. By resolving the audience's central objection first, the argument becomes substantially more persuasive to the intended audience, demonstrating how explicit audience modeling can guide both rhetorical planning and the final argument generation.

\section{Conclusion}
\label{sec:conclusion}
We presented \textsc{Argus}, an agent-based framework that advances computational argumentation along three connected axes: an explicit ToM reasoner that externalizes the audience's beliefs and values before writing, a component-aware planner that assigns rhetorical functions and grounds each subtopic in evidence at planning time, and a multi-dimensional refiner that targets weaknesses without quality regression. Across three benchmarks and multiple backbones, \textsc{Argus} consistently outperforms strong baselines, and our targeted simulations show these gains translate into genuine stance shifts in audiences rather than mere fluency. 

\section*{Limitations}

Several limitations remain. First, the Theory-of-Mind model is itself produced by an LLM and may not faithfully capture real audience mental states; it encodes a plausible model of the audience rather than a verified one. Second, despite planning-time evidence retrieval, individual sections can still contain inaccurate claims, and our error study shows the sharpest failure mode: when the assigned stance requires arguing against empirical consensus, no amount of refinement can manufacture grounding the evidence does not support. A dedicated fact-checking agent is a natural next step. Third, the multi-stage pipeline adds latency over single-pass generation, which parallelizing independent agents could mitigate.

Another question is how to evaluate persuasion, and we
approach it from complementary angles rather than trusting any single signal. Our human study measures overall preference directly, but such judgments might be confounded by annotators' prior beliefs on controversial propositions: a ``persuasive'' argument is too easily conflated with one the annotator already agrees with. Our targeted simulation mitigates this by fixing the audience's starting position, where a judge role-plays a skeptical reader holding the original stance, yielding a more controlled and reproducible proxy. Each lens is imperfect in its own way: human preference carries belief-driven bias, whereas
LLM judges may harbor biases of their own and can favor model-generated text. We therefore treat \emph{agreement across human and automatic protocols}, rather than any one metric, as our primary evidence, and leave validating simulated stance-shift against measured human stance-shift to future work.

\section*{Ethical Considerations}

Persuasive technology is inherently dual-use: the same audience modeling that makes an argument resonate can also enable manipulation, disinformation, or influence operations. The risk is heightened precisely because our framework conditions on a target's beliefs and emotional triggers. We therefore advocate clear disclosure of AI-generated arguments, deployment guidelines that prohibit deceptive use, and continued research on argument provenance and detection. We intend this work to advance the understanding of computational argumentation and to support legitimate applications such as debate education, writing assistance, and policy analysis.

\nocite{*}

\bibliographystyle{unsrtnat}
\bibliography{custom}

@inproceedings{ding-etal-2025-feat,
    title = "{FEAT}-writing: An Interactive Training System for Argumentative Writing",
    author = "Ding, Yuning  and
      Wehrhahn, Franziska  and
      Horbach, Andrea",
    editor = "Rambow, Owen  and
      Wanner, Leo  and
      Apidianaki, Marianna  and
      Al-Khalifa, Hend  and
      Eugenio, Barbara Di  and
      Schockaert, Steven  and
      Mather, Brodie  and
      Dras, Mark",
    booktitle = "Proceedings of the 31st International Conference on Computational Linguistics: System Demonstrations",
    month = jan,
    year = "2025",
    address = "Abu Dhabi, UAE",
    publisher = "Association for Computational Linguistics",
    url = "https://aclanthology.org/2025.coling-demos.22/",
    pages = "217--225"
}

@inproceedings{bao-etal-2022-aeg,
    title = "{AEG}: Argumentative Essay Generation via A Dual-Decoder Model with Content Planning",
    author = "Bao, Jianzhu  and
      Wang, Yasheng  and
      Li, Yitong  and
      Mi, Fei  and
      Xu, Ruifeng",
    editor = "Goldberg, Yoav  and
      Kozareva, Zornitsa  and
      Zhang, Yue",
    booktitle = "Proceedings of the 2022 Conference on Empirical Methods in Natural Language Processing",
    month = dec,
    year = "2022",
    address = "Abu Dhabi, United Arab Emirates",
    publisher = "Association for Computational Linguistics",
    url = "https://aclanthology.org/2022.emnlp-main.343/",
    doi = "10.18653/v1/2022.emnlp-main.343",
    pages = "5134--5148"
}

@article{slonim2021autonomous,
  title={An autonomous debating system},
  author={Slonim, Noam and Bilu, Yonatan and Alzate, Carlos and Bar-Haim, Roy and Bogin, Ben and Bonin, Francesca and Choshen, Leshem and Cohen-Karlik, Edo and Dankin, Lena and Edelstein, Lilach and others},
  journal={Nature},
  volume={591},
  number={7850},
  pages={379--384},
  year={2021},
  publisher={Nature Publishing Group UK London}
}

@inproceedings{florou-etal-2013-argument,
    title = "Argument extraction for supporting public policy formulation",
    author = "Florou, Eirini  and
      Konstantopoulos, Stasinos  and
      Koukourikos, Antonis  and
      Karampiperis, Pythagoras",
    editor = "Lendvai, Piroska  and
      Zervanou, Kalliopi",
    booktitle = "Proceedings of the 7th Workshop on Language Technology for Cultural Heritage, Social Sciences, and Humanities",
    month = aug,
    year = "2013",
    address = "Sofia, Bulgaria",
    publisher = "Association for Computational Linguistics",
    url = "https://aclanthology.org/W13-2707/",
    pages = "49--54"
}

@inproceedings{verma-etal-2026-predicting,
    title = "Predicting Convincingness in Political Speech: How Emotional Tone Shapes Persuasive Strength",
    author = "Verma, Bhuvanesh  and
      Marreddy, Mounika  and
      Mehler, Alexander",
    editor = "Barnes, Jeremy  and
      Barriere, Valentin  and
      De Clercq, Orph{\'e}e  and
      Klinger, Roman  and
      Nouri, C{\'e}lia  and
      Nozza, Debora  and
      Singh, Pranaydeep",
    booktitle = "The Proceedings for the 15th Workshop on Computational Approaches to Subjectivity, Sentiment Social Media Analysis ({WASSA} 2026)",
    month = mar,
    year = "2026",
    address = "Rabat, Morocco",
    publisher = "Association for Computational Linguistics",
    url = "https://aclanthology.org/2026.wassa-1.4/",
    doi = "10.18653/v1/2026.wassa-1.4",
    pages = "37--51",
    ISBN = "979-8-89176-378-4"
}

@inproceedings{wang-etal-2019-persuasion,
    title = "Persuasion for Good: Towards a Personalized Persuasive Dialogue System for Social Good",
    author = "Wang, Xuewei  and
      Shi, Weiyan  and
      Kim, Richard  and
      Oh, Yoojung  and
      Yang, Sijia  and
      Zhang, Jingwen  and
      Yu, Zhou",
    editor = "Korhonen, Anna  and
      Traum, David  and
      M{\`a}rquez, Llu{\'i}s",
    booktitle = "Proceedings of the 57th Annual Meeting of the Association for Computational Linguistics",
    month = jul,
    year = "2019",
    address = "Florence, Italy",
    publisher = "Association for Computational Linguistics",
    url = "https://aclanthology.org/P19-1566/",
    doi = "10.18653/v1/P19-1566",
    pages = "5635--5649"
}

@inproceedings{priya-etal-2025-argue,
    title = "We Argue to Agree: Towards Personality-Driven Argumentation-Based Negotiation Dialogue Systems for Tourism",
    author = "Priya, Priyanshu  and
      Dudhate, Saurav  and
      Yasheshbhai, Desai Vishesh  and
      Ekbal, Asif",
    editor = "Christodoulopoulos, Christos  and
      Chakraborty, Tanmoy  and
      Rose, Carolyn  and
      Peng, Violet",
    booktitle = "Findings of the Association for Computational Linguistics: EMNLP 2025",
    month = nov,
    year = "2025",
    address = "Suzhou, China",
    publisher = "Association for Computational Linguistics",
    url = "https://aclanthology.org/2025.findings-emnlp.1390/",
    doi = "10.18653/v1/2025.findings-emnlp.1390",
    pages = "25504--25536",
    ISBN = "979-8-89176-335-7"
}

@article{dyachenko2025llm,
  title={LLM services in the management of social communications},
  author={Dyachenko, Yuriy and Humenna, Oleksandra and Soloviov, Oleg and Skarga-Bandurova, Inna and Nenkov, Nayden},
  journal={Frontiers in Artificial Intelligence},
  volume={8},
  pages={1474017},
  year={2025},
  publisher={Frontiers Media SA}
}

@inproceedings{hu-etal-2025-debate,
    title = "Debate-to-Write: A Persona-Driven Multi-Agent Framework for Diverse Argument Generation",
    author = "Hu, Zhe  and
      Chan, Hou Pong  and
      Li, Jing  and
      Yin, Yu",
    editor = "Rambow, Owen  and
      Wanner, Leo  and
      Apidianaki, Marianna  and
      Al-Khalifa, Hend  and
      Eugenio, Barbara Di  and
      Schockaert, Steven",
    booktitle = "Proceedings of the 31st International Conference on Computational Linguistics",
    month = jan,
    year = "2025",
    address = "Abu Dhabi, UAE",
    publisher = "Association for Computational Linguistics",
    url = "https://aclanthology.org/2025.coling-main.314/",
    pages = "4689--4703"
}

@article{han2026drpg,
  title={DRPG (Decompose, Retrieve, Plan, Generate): An Agentic Framework for Academic Rebuttal},
  author={Han, Peixuan and Yu, Yingjie and Xu, Jingjun and You, Jiaxuan},
  journal={arXiv preprint arXiv:2601.18081},
  year={2026}
}

@inproceedings{zhao-etal-2025-plan,
    title = "Plan Dynamically, Express Rhetorically: A Debate-Driven Rhetorical Framework for Argumentative Writing",
    author = "Zhao, Xueguan  and
      Lu, Wenpeng  and
      Zheng, Chaoqun  and
      Zhang, Weiyu  and
      Si, Jiasheng  and
      Zhou, Deyu",
    editor = "Christodoulopoulos, Christos  and
      Chakraborty, Tanmoy  and
      Rose, Carolyn  and
      Peng, Violet",
    booktitle = "Proceedings of the 2025 Conference on Empirical Methods in Natural Language Processing",
    month = nov,
    year = "2025",
    address = "Suzhou, China",
    publisher = "Association for Computational Linguistics",
    url = "https://aclanthology.org/2025.emnlp-main.483/",
    doi = "10.18653/v1/2025.emnlp-main.483",
    pages = "9551--9573",
    ISBN = "979-8-89176-332-6"
}

@inproceedings{hu-etal-2024-americano-argument,
    title = "{AMERICANO}: Argument Generation with Discourse-driven Decomposition and Agent Interaction",
    author = "Hu, Zhe  and
      Chan, Hou Pong  and
      Yin, Yu",
    editor = "Mahamood, Saad  and
      Minh, Nguyen Le  and
      Ippolito, Daphne",
    booktitle = "Proceedings of the 17th International Natural Language Generation Conference",
    month = sep,
    year = "2024",
    address = "Tokyo, Japan",
    publisher = "Association for Computational Linguistics",
    url = "https://aclanthology.org/2024.inlg-main.8/",
    doi = "10.18653/v1/2024.inlg-main.8",
    pages = "82--102"
}

@inproceedings{xiao-etal-2024-prove,
    title = "Prove Your Point!: Bringing Proof-Enhancement Principles to Argumentative Essay Generation",
    author = "Xiao, Ruiyu  and
      Wu, Lei  and
      Gou, Yuhang  and
      Zhang, Weinan  and
      Liu, Ting",
    editor = "Al-Onaizan, Yaser  and
      Bansal, Mohit  and
      Chen, Yun-Nung",
    booktitle = "Proceedings of the 2024 Conference on Empirical Methods in Natural Language Processing",
    month = nov,
    year = "2024",
    address = "Miami, Florida, USA",
    publisher = "Association for Computational Linguistics",
    url = "https://aclanthology.org/2024.emnlp-main.1058/",
    doi = "10.18653/v1/2024.emnlp-main.1058",
    pages = "18995--19008"
}

@inproceedings{yeginbergen-etal-2025-dynamic,
    title = "Dynamic Knowledge Integration for Evidence-Driven Counter-Argument Generation with Large Language Models",
    author = "Yeginbergen, Anar  and
      Oronoz, Maite  and
      Agerri, Rodrigo",
    editor = "Che, Wanxiang  and
      Nabende, Joyce  and
      Shutova, Ekaterina  and
      Pilehvar, Mohammad Taher",
    booktitle = "Findings of the Association for Computational Linguistics: ACL 2025",
    month = jul,
    year = "2025",
    address = "Vienna, Austria",
    publisher = "Association for Computational Linguistics",
    url = "https://aclanthology.org/2025.findings-acl.1161/",
    doi = "10.18653/v1/2025.findings-acl.1161",
    pages = "22568--22584",
    ISBN = "979-8-89176-256-5"
}

@article{li2025r,
  title={R-Debater: Retrieval-Augmented Debate Generation through Argumentative Memory},
  author={Li, Maoyuan and Wang, Zhongsheng and Li, Haoyuan and Liu, Jiamou},
  journal={arXiv preprint arXiv:2512.24684},
  year={2025}
}

@inproceedings{hua-etal-2019-argument-generation,
    title = "Argument Generation with Retrieval, Planning, and Realization",
    author = "Hua, Xinyu  and
      Hu, Zhe  and
      Wang, Lu",
    editor = "Korhonen, Anna  and
      Traum, David  and
      M{\`a}rquez, Llu{\'i}s",
    booktitle = "Proceedings of the 57th Annual Meeting of the Association for Computational Linguistics",
    month = jul,
    year = "2019",
    address = "Florence, Italy",
    publisher = "Association for Computational Linguistics",
    url = "https://aclanthology.org/P19-1255/",
    doi = "10.18653/v1/P19-1255",
    pages = "2661--2672"
}

@inproceedings{hu-etal-2022-planet,
    title = "{PLANET}: Dynamic Content Planning in Autoregressive Transformers for Long-form Text Generation",
    author = "Hu, Zhe  and
      Chan, Hou Pong  and
      Liu, Jiachen  and
      Xiao, Xinyan  and
      Wu, Hua  and
      Huang, Lifu",
    editor = "Muresan, Smaranda  and
      Nakov, Preslav  and
      Villavicencio, Aline",
    booktitle = "Proceedings of the 60th Annual Meeting of the Association for Computational Linguistics (Volume 1: Long Papers)",
    month = may,
    year = "2022",
    address = "Dublin, Ireland",
    publisher = "Association for Computational Linguistics",
    url = "https://aclanthology.org/2022.acl-long.163/",
    doi = "10.18653/v1/2022.acl-long.163",
    pages = "2288--2305"
}

@inproceedings{wang2023argument,
  title={Argument and counter-argument generation: A critical survey},
  author={Wang, Xiaoou and Cabrio, Elena and Villata, Serena},
  booktitle={International conference on applications of natural language to information systems},
  pages={500--510},
  year={2023},
  organization={Springer}
}

@article{sichach2024ethos,
  title={Ethos, pathos and logos as foundations of persuasive writing},
  author={Sichach, Moses},
  journal={Available at SSRN 4971293},
  year={2024}
}

@inproceedings{higgins2012ethos,
  title={Ethos, logos, pathos: Strategies of persuasion in social/environmental reports},
  author={Higgins, Colin and Walker, Robyn},
  booktitle={Accounting forum},
  volume={36},
  number={3},
  pages={194--208},
  year={2012},
  organization={Elsevier}
}

@article{wang-etal-2017-winning,
    title = "Winning on the Merits: The Joint Effects of Content and Style on Debate Outcomes",
    author = "Wang, Lu  and
      Beauchamp, Nick  and
      Shugars, Sarah  and
      Qin, Kechen",
    editor = "Lee, Lillian  and
      Johnson, Mark  and
      Toutanova, Kristina",
    journal = "Transactions of the Association for Computational Linguistics",
    volume = "5",
    year = "2017",
    address = "Cambridge, MA",
    publisher = "MIT Press",
    url = "https://aclanthology.org/Q17-1016/",
    doi = "10.1162/tacl_a_00057",
    pages = "219--232"
}

@book{seo2023good,
  title={Good arguments: How debate teaches us to listen and be heard},
  author={Seo, Bo},
  year={2023},
  publisher={Penguin}
}

@article{deane2015key,
  title={The key practice, discuss and debate ideas: Conceptual framework, literature review, and provisional learning progressions for argumentation},
  author={Deane, Paul and Song, Yi},
  journal={ETS Research Report Series},
  volume={2015},
  number={2},
  pages={1--21},
  year={2015},
  publisher={Wiley Online Library}
}

@book{andriessen1999planning,
  title={From planning to translating: The specificity of argumentative writing.},
  author={Andriessen, JEB and Chanquoy, Lucile and Coirier, Pierre and others},
  year={1999},
  publisher={Amsterdam University Press}
}

@inproceedings{sorensen2024value,
  title={Value kaleidoscope: Engaging ai with pluralistic human values, rights, and duties},
  author={Sorensen, Taylor and Jiang, Liwei and Hwang, Jena D and Levine, Sydney and Pyatkin, Valentina and West, Peter and Dziri, Nouha and Lu, Ximing and Rao, Kavel and Bhagavatula, Chandra and others},
  booktitle={Proceedings of the AAAI Conference on Artificial Intelligence},
  volume={38},
  number={18},
  pages={19937--19947},
  year={2024}
}

@inproceedings{hua-etal-2021-dyploc,
    title = "{DYPLOC}: Dynamic Planning of Content Using Mixed Language Models for Text Generation",
    author = "Hua, Xinyu  and
      Sreevatsa, Ashwin  and
      Wang, Lu",
    editor = "Zong, Chengqing  and
      Xia, Fei  and
      Li, Wenjie  and
      Navigli, Roberto",
    booktitle = "Proceedings of the 59th Annual Meeting of the Association for Computational Linguistics and the 11th International Joint Conference on Natural Language Processing (Volume 1: Long Papers)",
    month = aug,
    year = "2021",
    address = "Online",
    publisher = "Association for Computational Linguistics",
    url = "https://aclanthology.org/2021.acl-long.501/",
    doi = "10.18653/v1/2021.acl-long.501",
    pages = "6408--6423"
}

@inproceedings{he-etal-2024-decomposing,
    title = "Decomposing Argumentative Essay Generation via Dialectical Planning of Complex Reasoning",
    author = "He, Yuhang  and
      Bao, Jianzhu  and
      Sun, Yang  and
      Liang, Bin  and
      Yang, Min  and
      Qin, Bing  and
      Xu, Ruifeng",
    editor = "Ku, Lun-Wei  and
      Martins, Andre  and
      Srikumar, Vivek",
    booktitle = "Findings of the Association for Computational Linguistics: ACL 2024",
    month = aug,
    year = "2024",
    address = "Bangkok, Thailand",
    publisher = "Association for Computational Linguistics",
    url = "https://aclanthology.org/2024.findings-acl.731/",
    doi = "10.18653/v1/2024.findings-acl.731",
    pages = "12305--12322"
}

@inproceedings{wachsmuth2018argumentation,
  title={Argumentation synthesis following rhetorical strategies},
  author={Wachsmuth, Henning and Stede, Manfred and El Baff, Roxanne and Al Khatib, Khalid and Skeppstedt, Maria and Stein, Benno},
  booktitle={Proceedings of the 27th international conference on computational linguistics},
  pages={3753--3765},
  year={2018}
}

@article{madaan2023self,
  title={Self-refine: Iterative refinement with self-feedback},
  author={Madaan, Aman and Tandon, Niket and Gupta, Prakhar and Hallinan, Skyler and Gao, Luyu and Wiegreffe, Sarah and Alon, Uri and Dziri, Nouha and Prabhumoye, Shrimai and Yang, Yiming and others},
  journal={Advances in neural information processing systems},
  volume={36},
  pages={46534--46594},
  year={2023}
}

@inproceedings{saha-etal-2021-explagraphs,
    title = "{E}xpla{G}raphs: An Explanation Graph Generation Task for Structured Commonsense Reasoning",
    author = "Saha, Swarnadeep  and
      Yadav, Prateek  and
      Bauer, Lisa  and
      Bansal, Mohit",
    editor = "Moens, Marie-Francine  and
      Huang, Xuanjing  and
      Specia, Lucia  and
      Yih, Scott Wen-tau",
    booktitle = "Proceedings of the 2021 Conference on Empirical Methods in Natural Language Processing",
    month = nov,
    year = "2021",
    address = "Online and Punta Cana, Dominican Republic",
    publisher = "Association for Computational Linguistics",
    url = "https://aclanthology.org/2021.emnlp-main.609/",
    doi = "10.18653/v1/2021.emnlp-main.609",
    pages = "7716--7740"
}

@article{elo1967proposed,
  title={The proposed uscf rating system, its development, theory, and applications},
  author={Elo, Arpad E},
  journal={Chess life},
  volume={22},
  number={8},
  pages={242--247},
  year={1967}
}

@article{bai2022training,
  title={Training a helpful and harmless assistant with reinforcement learning from human feedback},
  author={Bai, Yuntao and Jones, Andy and Ndousse, Kamal and Askell, Amanda and Chen, Anna and DasSarma, Nova and Drain, Dawn and Fort, Stanislav and Ganguli, Deep and Henighan, Tom and others},
  journal={arXiv preprint arXiv:2204.05862},
  year={2022}
}

@inproceedings{hua-wang-2018-neural,
    title = "Neural Argument Generation Augmented with Externally Retrieved Evidence",
    author = "Hua, Xinyu  and
      Wang, Lu",
    editor = "Gurevych, Iryna  and
      Miyao, Yusuke",
    booktitle = "Proceedings of the 56th Annual Meeting of the Association for Computational Linguistics (Volume 1: Long Papers)",
    month = jul,
    year = "2018",
    address = "Melbourne, Australia",
    publisher = "Association for Computational Linguistics",
    url = "https://aclanthology.org/P18-1021/",
    doi = "10.18653/v1/P18-1021",
    pages = "219--230"
}

@inproceedings{hua2021dyploc,
  title={DYPLOC: Dynamic planning of content using mixed language models for text generation},
  author={Hua, Xinyu and Sreevatsa, Ashwin and Wang, Lu},
  booktitle={Proceedings of the 59th Annual Meeting of the Association for Computational Linguistics and the 11th International Joint Conference on Natural Language Processing (Volume 1: Long Papers)},
  pages={6408--6423},
  year={2021}
}

@inproceedings{schiller2021aspect,
  title={Aspect-controlled neural argument generation},
  author={Schiller, Benjamin and Daxenberger, Johannes and Gurevych, Iryna},
  booktitle={Proceedings of the 2021 Conference of the North American Chapter of the Association for Computational Linguistics: Human Language Technologies},
  pages={380--396},
  year={2021}
}

@article{li2025large,
  title={Large language models in argument mining: A survey},
  author={Li, Hao and Schlegel, Viktor and Sun, Yizheng and Batista-Navarro, Riza and Nenadic, Goran},
  journal={arXiv preprint arXiv:2506.16383},
  year={2025}
}

@article{wei2022chain,
  title={Chain-of-thought prompting elicits reasoning in large language models},
  author={Wei, Jason and Wang, Xuezhi and Schuurmans, Dale and Bosma, Maarten and Xia, Fei and Chi, Ed and Le, Quoc V and Zhou, Denny and others},
  journal={Advances in neural information processing systems},
  volume={35},
  pages={24824--24837},
  year={2022}
}

@article{nguyen2025survey,
  title={A survey of theory of mind in large language models: Evaluations, representations, and safety risks},
  author={Nguyen, Hieu Minh and others},
  journal={arXiv preprint arXiv:2502.06470},
  year={2025}
}

@article{moore2025large,
  title={Do Large Language Models Have a Planning Theory of Mind? Evidence from MINDGAMES: a Multi-Step Persuasion Task},
  author={Moore, Jared and Cooper, Ned and Overmark, Rasmus and Cibralic, Beba and Haber, Nick and Jones, Cameron R},
  journal={arXiv preprint arXiv:2507.16196},
  year={2025}
}

@inproceedings{singh2025measuring,
  title={Measuring and improving persuasiveness of large language models},
  author={Singh, Somesh and Singla, Yaman and Si, Harini and Krishnamurthy, Balaji},
  booktitle={International Conference on Learning Representations},
  volume={2025},
  pages={90267--90322},
  year={2025}
}

@online{durmus2024persuasion,
author = {Esin Durmus and Liane Lovitt and Alex Tamkin and Stuart Ritchie and Jack Clark and Deep Ganguli},
title = {Measuring the Persuasiveness of Language Models},
date = {2024-04-09},
year = {2024},
url = {https://www.anthropic.com/news/measuring-model-persuasiveness},
}

@article{hinton2023persuasive,
  title={How persuasive is AI-generated argumentation? An analysis of the quality of an argumentative text produced by the GPT-3 AI text generator},
  author={Hinton, Martin and Wagemans, Jean HM},
  journal={Argument \& Computation},
  volume={14},
  number={1},
  pages={59--74},
  year={2023},
  publisher={SAGE Publications Sage UK: London, England}
}

@article{gleason1999role,
  title={The role of evidence in argumentative writing},
  author={Gleason, Mary M},
  journal={Reading \& Writing Quarterly},
  volume={15},
  number={1},
  pages={81--106},
  year={1999},
  publisher={Taylor \& Francis}
}

@incollection{jung2026argumentation,
  title={Argumentation},
  author={Jung, Nathan},
  booktitle={The Process of Generative AI Writing: A Practical Guide for Undergraduates Across Disciplines},
  pages={83--96},
  year={2026},
  publisher={Springer}
}

@inproceedings{chen-etal-2024-complex,
    title = "Complex Claim Verification with Evidence Retrieved in the Wild",
    author = "Chen, Jifan  and
      Kim, Grace  and
      Sriram, Aniruddh  and
      Durrett, Greg  and
      Choi, Eunsol",
    editor = "Duh, Kevin  and
      Gomez, Helena  and
      Bethard, Steven",
    booktitle = "Proceedings of the 2024 Conference of the North American Chapter of the Association for Computational Linguistics: Human Language Technologies (Volume 1: Long Papers)",
    month = jun,
    year = "2024",
    address = "Mexico City, Mexico",
    publisher = "Association for Computational Linguistics",
    url = "https://aclanthology.org/2024.naacl-long.196/",
    doi = "10.18653/v1/2024.naacl-long.196",
    pages = "3569--3587"
}

@inproceedings{chen-etal-2024-humans,
    title = "Humans or {LLM}s as the Judge? A Study on Judgement Bias",
    author = "Chen, Guiming Hardy  and
      Chen, Shunian  and
      Liu, Ziche  and
      Jiang, Feng  and
      Wang, Benyou",
    editor = "Al-Onaizan, Yaser  and
      Bansal, Mohit  and
      Chen, Yun-Nung",
    booktitle = "Proceedings of the 2024 Conference on Empirical Methods in Natural Language Processing",
    month = nov,
    year = "2024",
    address = "Miami, Florida, USA",
    publisher = "Association for Computational Linguistics",
    url = "https://aclanthology.org/2024.emnlp-main.474/",
    doi = "10.18653/v1/2024.emnlp-main.474",
    pages = "8301--8327"
}

@article{wu2026comparing,
  title={Comparing GPT and human raters in essay assessment: Variability, bias, and the potential of LLM-based scoring},
  author={Wu, Hsiang-Ning and Chu, Man-Ni and Hsu, Jia-Lien},
  journal={Computers and Education Open},
  pages={100341},
  year={2026},
  publisher={Elsevier}
}

@article{han2025tomap,
      title={ToMAP: Training Opponent-Aware LLM Persuaders with Theory of Mind}, 
      author={Peixuan Han and Zijia Liu and Jiaxuan You},
      year={2025},
      journal={arXiv preprint arXiv:2505.22961},
      archivePrefix={arXiv},
      url={https://arxiv.org/abs/2505.22961}, 
}

@article{moore2026large,
  title={Large Language Models Persuade Without Planning Theory of Mind},
  author={Moore, Jared and Overmark, Rasmus and Cooper, Ned and Cibralic, Beba and Haber, Nick and Jones, Cameron R},
  journal={arXiv preprint arXiv:2602.17045},
  year={2026}
}

@inproceedings{falk2023storyarg,
  title={StoryARG: a corpus of narratives and personal experiences in argumentative texts},
  author={Falk, Neele and Lapesa, Gabriella},
  booktitle={Proceedings of the 61st Annual Meeting of the Association for Computational Linguistics (Volume 1: Long Papers)},
  pages={2350--2372},
  year={2023}
}

@article{fyu2025persuasivetom,
  title={Persuasivetom: A benchmark for evaluating machine theory of mind in persuasive dialogues},
  author={Yu, Fangxu and Jiang, Lai and Huang, Shenyi and Wu, Zhen and Dai, Xinyu},
  journal={arXiv preprint arXiv:2502.21017},
  year={2025}
}

@article{ma2026think,
  title={Think Thrice Before You Speak: Dual knowledge-enhanced Theory-of-Mind Reasoning for Persuasive Agents},
  author={Ma, Minghui and Guo, Bin and Yang, Runze and Chen, Mengqi and Liu, Yan and Liu, Jingqi and Pei, Yahan and Ma, Xuehao and Zhang, Qiuyun and Yu, Zhiwen},
  journal={arXiv preprint arXiv:2605.22602},
  year={2026}
}

@misc{zhang2026ma2pmetacognitiveautonomousintelligent,
      title={MA$^{2}$P: A Meta-Cognitive Autonomous Intelligent Agents Framework for Complex Persuasion}, 
      author={Dingyi Zhang and Ziqing Zhuang and Linhai Zhang and Ziyang Gao and Deyu Zhou},
      year={2026},
      eprint={2605.18572},
      archivePrefix={arXiv},
      primaryClass={cs.CL},
      url={https://arxiv.org/abs/2605.18572}, 
}

@inproceedings{jin-etal-2026-arggenbench,
    title = "{A}rg{G}en{B}ench: Benchmarking the Complex Controlled Argument Generation Capability of Large Language Models",
    author = "Jin, Bojun  and
      Bao, Jianzhu  and
      Sun, Yang  and
      Zhang, Yice  and
      Xu, Ruifeng",
    editor = "Liakata, Maria  and
      Moreira, Viviane P.  and
      Zhang, Jiajun  and
      Jurgens, David",
    booktitle = "Proceedings of the 64th Annual Meeting of the {A}ssociation for {C}omputational {L}inguistics (Volume 1: Long Papers)",
    month = jul,
    year = "2026",
    address = "San Diego, California, United States",
    publisher = "Association for Computational Linguistics",
    url = "https://aclanthology.org/2026.acl-long.1414/",
    doi = "10.18653/v1/2026.acl-long.1414",
    pages = "30630--30662",
    ISBN = "979-8-89176-390-6"
}


\clearpage
\begin{appendix}
    
\appendix

\section{Implementation Details}
\label{app:details}


\paragraph{Backbone LLM and evaluation judge.}
Three instruction-tuned backbone LLMs are utilized as backbone model for implementations: {DeepSeek-v3.2},
{Qwen3.5-Flash-2026-02-23}, and {gpt-5-mini-2025-08-07}. The
\textbf{evaluation judge is fixed to GPT-5.4} across all conditions,
independent of the generation backbone. We leverage the official API call for model implementation.

During generation, all JSON outputs use up to 3 retries with exponential backoff. Both the maximum planning iteration and refinement rounds are set as 2. For WebSearch, we leverage the ddgs library~\footnote{\url{https://github.com/deedy5/ddgs}} to retrieve URLs and short snippets, and utilize trafilatura~\footnote{\url{https://github.com/adbar/trafilatura}} to parse the webpage. 

We evaluate our methods on three datasets including Reddit/CMV, iDebate, and ExplaGraph.
For each dataset, we follow previous work~\cite{hu-etal-2025-debate} and randomly sample 30 inputs for evaluation.

\begin{figure}[h]
\begin{tcolorbox}[
  colback=gray!5, colframe=blue!50, arc=3pt,
  title={\small\textbf{Input Analysis: Example Output}},
  fonttitle=\small, left=5pt, right=5pt, top=3pt, bottom=3pt
]
\small
\textbf{Proposition:} \textit{``Hate speech is free speech.''}\\[3pt]
\textbf{Background:}\footnote{The \textit{background} field is an LLM summary grounded in retrieved snippets; \textit{retrieved evidence} lists the specific factual claims extracted from those snippets.} The claim equates hate speech with constitutionally
protected expression. U.S.\ First Amendment doctrine recognizes no formal
``hate speech'' exception, though narrow exclusions apply (incitement,
true threats, fighting words). Empirical work links hate speech exposure
to measurable psychological harm.\\[3pt]
\textbf{Key claims extracted:}
\begin{itemize}[nosep,leftmargin=1.4em,itemsep=1pt]
  \item Hate speech is legally subsumed under free speech protections.
  \item Restricting hate speech constitutes impermissible censorship.
\end{itemize}
\vspace{2pt}
\textbf{Retrieved evidence (grounded from web retrieval):}
\begin{itemize}[nosep,leftmargin=1.4em,itemsep=1pt]
  \item No general hate speech exception exists under U.S.\ law
        (\textit{Matal v.~Tam}, 2017; \textit{R.A.V.\ v.\ City of
        St.\ Paul}, 1992).
  \item Hate speech exposure correlates with increased anxiety,
        depression, and reduced civic participation among targeted groups.
  \item Germany, the UK, and France criminalize incitement to hatred,
        demonstrating that the legal equivalence is contestable across
        democratic systems.
\end{itemize}
\end{tcolorbox}
\caption{Structured output of the Input Analyzer for a  proposition. }
\label{fig:input-analysis-example}
\end{figure}
\begin{figure}[t]
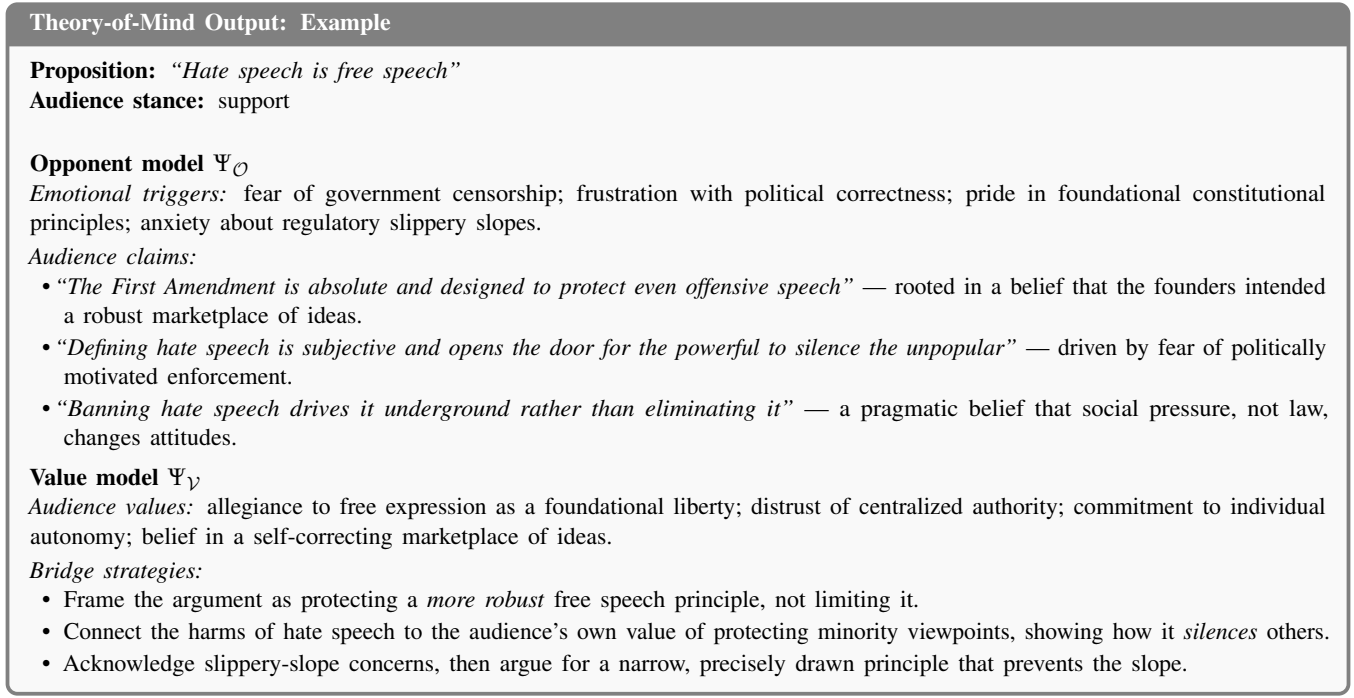

\begin{tcolorbox}[
  colback=gray!5, colframe=black!50, arc=3pt,
  title={\small\textbf{Theory-of-Mind Output: Example}},
  fonttitle=\small, left=5pt, right=5pt, top=3pt, bottom=3pt
]
\small
\textbf{Proposition:} \textit{``Hate speech is free speech''} \quad\\
\textbf{Audience stance:} support\\[2pt]

\textbf{Opponent model $\Psi_\mathcal{O}$}\\
\textit{Emotional triggers:} fear of government censorship; frustration with
political correctness; pride in foundational constitutional principles; anxiety
about regulatory slippery slopes.\\[2pt]
\textit{Audience claims:}
\begin{itemize}[nosep,leftmargin=1.4em,itemsep=1pt]
  \item \textit{``The First Amendment is absolute and designed to protect even
    offensive speech''} --- rooted in a belief that the founders intended a
    robust marketplace of ideas.
  \item \textit{``Defining hate speech is subjective and opens the door for
    the powerful to silence the unpopular''} --- driven by fear of politically
    motivated enforcement.
  \item \textit{``Banning hate speech drives it underground rather than
    eliminating it''} --- a pragmatic belief that social pressure, not law,
    changes attitudes.
\end{itemize}
\vspace{4pt}

\textbf{Value model $\Psi_\mathcal{V}$}\\
\textit{Audience values:} allegiance to free expression as a foundational
liberty; distrust of centralized authority; commitment to individual autonomy;
belief in a self-correcting marketplace of ideas.\\[2pt]
\textit{Bridge strategies:}
\begin{itemize}[nosep,leftmargin=1.4em,itemsep=1pt]
  \item Frame the argument as protecting a \emph{more robust} free speech
    principle, not limiting it.
  \item Connect the harms of hate speech to the audience's own value of
    protecting minority viewpoints, showing how it \emph{silences} others.
  \item Acknowledge slippery-slope concerns, then argue for a narrow,
    precisely drawn principle that prevents the slope.
\end{itemize}
\end{tcolorbox}
\caption{ToM Reasoner output $\Psi$ for a proposition.}
\label{fig:tom-example}
\end{figure}

\paragraph{Baselines.}
All four baselines use the same backbone LLM, and
operate zero-shot prompting as our model.  We keep all parameters the same as ours during the inference.

\section{\textsc{Argus} Module Prompts}
\label{app:prompts}

\begin{figure}[h]
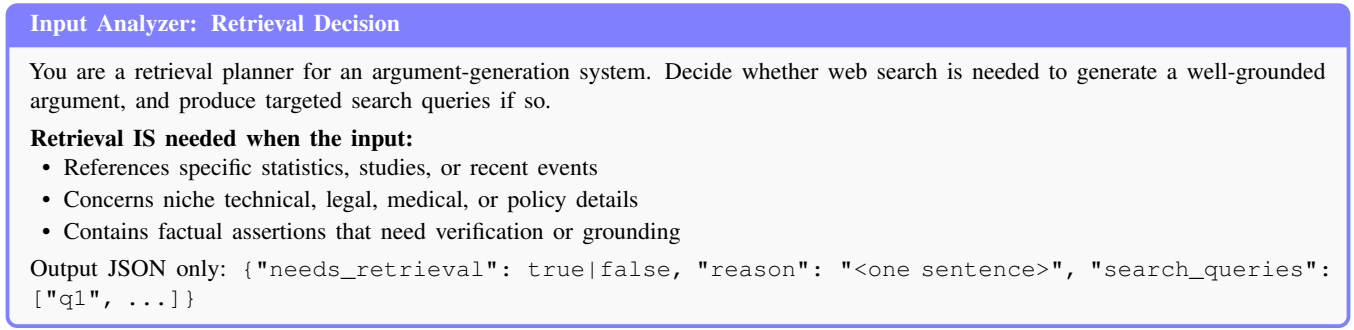

\begin{tcolorbox}[
  colback=gray!5, colframe=blue!50, arc=3pt,
  title={\small\textbf{Input Analyzer: Retrieval Decision}},
  fonttitle=\small, left=5pt, right=5pt, top=3pt, bottom=3pt
]
\small
You are a retrieval planner for an argument-generation system. Decide
whether web search is needed to generate a well-grounded argument, and
produce targeted search queries if so.\\[3pt]
\textbf{Retrieval IS needed when the input:}
\begin{itemize}[nosep,leftmargin=1.4em,itemsep=1pt]
  \item References specific statistics, studies, or recent events
  \item Concerns niche technical, legal, medical, or policy details
  \item Contains factual assertions that need verification or grounding
\end{itemize}
\vspace{3pt}
Output JSON only: \texttt{\{"needs\_retrieval": true|false, "reason":
"<one sentence>", "search\_queries": ["q1", \ldots]\}}
\end{tcolorbox}
\caption{Input Analyzer system prompt. A fast LLM call decides
  whether retrieval is warranted and generates 2--4 targeted queries,
  which are executed \emph{before} analysis so Phase~3 is grounded in
  actual retrieved content.}
\label{fig:prompt-ia1}
\end{figure}

Figures~\ref{fig:prompt-ia1}--\ref{fig:prompt-elo} report the
system prompts for all \textsc{Argus} modules and evaluators. Red-bordered boxes
mark the three novel components ($\star$). User message templates are shown where
informative; brackets \texttt{[\ldots]} denote dynamically inserted content.

\begin{figure}[t]
\begin{tcolorbox}[
  colback=gray!5, colframe=blue!50, arc=3pt,
  title={\small\textbf{Input Analyzer: Unified Analysis}},
  fonttitle=\small, left=5pt, right=5pt, top=3pt, bottom=3pt
]
\small
You are an expert argumentation analyst. Given an input statement and
optional retrieved context, analyze the argumentative dimensions of the
input for later argument planning and generation.\\[3pt]
\textbf{Extract} \texttt{key\_claims} stated explicitly in the input only
--- do not invent sub-dimensions. If retrieved context was provided,
synthesize \texttt{key\_evidence}: 2--5 concrete, usable factual claims
from the snippets (e.g., ``Studies show X\% of Y do Z''). If the input
is a full argument (not a short proposition), extract
\texttt{logical\_structure} with \textit{premises},
\textit{implicit\_assumptions}, and \textit{logical\_gaps}; omit
otherwise.\\[3pt]
Output JSON with fields:
\texttt{background\_context} (2--3 sentences of relevant context),
\texttt{key\_claims}, \texttt{key\_evidence}, \texttt{logical\_structure}.
\end{tcolorbox}
\caption{Input Analyzer system prompt. The unified analysis is
  grounded in any content retrieved in Phase~2. Its outputs
  $(\mathcal{K}, \mathcal{L}, \mathcal{B})$ are passed verbatim to the
  ToM Reasoner and Argument Planner.}
\label{fig:prompt-ia3}
\end{figure}

\begin{figure*}[t]
\begin{tcolorbox}[
  colback=red!3, colframe=red!60, arc=3pt,
  title={\small\textbf{Theory of Mind Reasoner --- System Prompt $\star$}},
  fonttitle=\small, left=5pt, right=5pt, top=3pt, bottom=3pt
]
\small
You are an expert in cognitive science, social psychology, and
argumentation theory. Your task is to model the mental states of a
TARGET AUDIENCE in a persuasive argument context.\\[3pt]
You will be given a proposition, the WRITER's stance, and the AUDIENCE's
stance (the opposing side). \textbf{This step is PURELY descriptive
mental modeling. Do NOT generate arguments or persuasive text.}\\[6pt]
\textbf{1.\ Audience Profile.} Model what the audience feels and claims:
\begin{itemize}[nosep,leftmargin=1.4em,itemsep=1pt]
  \item \texttt{emotional\_triggers}: emotional themes that resonate
    strongly with this audience (fears, hopes, frustrations, identities)
  \item \texttt{audience\_claims}: up to 5 positions the audience holds;
    for each provide \texttt{"claim"} and \texttt{"basis"} (the
    underlying reason, emotion, or belief driving it)
\end{itemize}
\vspace{4pt}
\textbf{2.\ Value Analysis.} Map the audience's value landscape:
\begin{itemize}[nosep,leftmargin=1.4em,itemsep=1pt]
  \item \texttt{audience\_values}: open-ended natural language phrases
    (e.g., ``desire for personal safety'', ``attachment to familiar ways
    of life'') --- do NOT classify as pro- or anti-argument; the Planner
    decides how to use them
  \item \texttt{bridge\_strategies}: specific ways to reframe the
    argument to align with their values or mitigate tensions
  \item \texttt{value\_framing\_summary}: short prose summary of the
    overall value landscape
\end{itemize}
\vspace{4pt}
Output a JSON object with top-level keys \texttt{"audience"} and
\texttt{"value\_analysis"} matching the structure above.\\[6pt]
\textit{User message template:}
\textbf{Proposition:} \texttt{[proposition]}\quad
\textbf{Writer's stance:} \texttt{[stance]}\quad
\textbf{Audience's stance:} \texttt{[opposite\_stance]}\\
\textbf{Background:} \texttt{[background\_context]}\quad
\textbf{Key claims / premises / gaps:} \texttt{[logical\_structure]}\\
\textbf{Retrieved background:} \texttt{[retrieved\_snippets]}\\
Perform the Theory of Mind analysis and return the JSON.
\end{tcolorbox}
\caption{ToM Reasoner system prompt.}
\label{fig:prompt-tom}
\end{figure*}

\begin{figure*}[t]
\begin{tcolorbox}[
  colback=red!3, colframe=red!60, arc=3pt,
  title={\small\textbf{Argument Planner --- Plan Generation System Prompt $\star$}},
  fonttitle=\small, left=5pt, right=5pt, top=3pt, bottom=3pt
]
\small
You are an expert argumentation strategist and professional writer.
Given a proposition, your stance, an input analysis, and a ToM analysis
of the target audience, create a comprehensive, strategically-sound
argument plan.\\[3pt]
\textbf{Plan requirements:}
\begin{itemize}[nosep,leftmargin=1.4em,itemsep=1pt]
  \item Derive a rhetorical strategy grounded in the audience's beliefs,
    values, biases, and likely resistance (from the ToM)
  \item Decompose the argument into 2--4 strategically coherent
    subtopics, each serving a distinct persuasive function
  \item Assign one \textit{primary component} (dominant persuasive mode)
    and optional secondary components to each subtopic
  \item Flag whether each subtopic requires external evidence
  \item For each subtopic, anticipate the most likely audience
    counterargument and provide value-alignment or reframing strategies
  \item Ensure subtopics collectively form a logically ordered and
    rhetorically effective flow
\end{itemize}
\vspace{4pt}
\textbf{Rhetorical components:} \textit{logos} (logical reasoning,
inference chains, syllogisms); \textit{pathos} (emotional appeal,
narrative, vivid examples); \textit{ethos} (credibility, expert
authority, shared values); \textit{evidence} (empirical data,
statistics, research citations).\\[4pt]
Output JSON with a \texttt{"subtopics"} array; each entry has:
\texttt{subtopic}, \texttt{primary\_component},
\texttt{secondary\_components}, \texttt{key\_points},
\texttt{evidence\_needed}, \texttt{audience\_rebuttal},
\texttt{value\_alignment\_notes}; plus top-level
\texttt{plan\_quality\_score} (0--10) and
\texttt{plan\_quality\_rationale}.
\end{tcolorbox}
\caption{Argument Planner plan-generation system prompt. 
  }
\label{fig:prompt-planner}
\end{figure*}

\begin{figure*}[h]
\begin{tcolorbox}[
  colback=gray!5, colframe=blue!50, arc=3pt,
  title={\small\textbf{Argument Writer --- System Prompt}},
  fonttitle=\small, left=5pt, right=5pt, top=3pt, bottom=3pt
]
\small
You are an expert persuasive writer. Given a proposition, a stance
(support or refute), a plan with subtopics and rhetorical components, and
insights about the audience's values and beliefs, write a coherent and
persuasive argument that follows the provided plan while maximizing
persuasive impact.\\[3pt]
\textbf{Instructions:}
\begin{itemize}[nosep,leftmargin=1.4em,itemsep=1pt]
  \item Develop each subtopic in continuous flowing prose --- no section
    headers or labeled sections
  \item Output ONLY the argument text (no headers, no JSON,
    no References section)
  \item You may reorganize the plan's rhetorical structure as needed
    to produce the most persuasive argument
\end{itemize}
\end{tcolorbox}
\caption{Argument Writer system prompt. 
  }
\label{fig:prompt-writer}
\end{figure*}

\begin{figure*}[t]
\begin{tcolorbox}[
  colback=gray!5, colframe=black!40, arc=3pt,
  title={\small\textbf{Quality Evaluator --- System Prompt (Absolute Scoring, GPT-5.4)}},
  fonttitle=\small, left=5pt, right=5pt, top=3pt, bottom=3pt
]
\small
You are an expert argument quality evaluator trained in rhetoric,
argumentation theory, and persuasion science. Assess arguments along
multiple dimensions and provide actionable feedback.\\[4pt]
\textbf{Scoring rubric:}
\begin{itemize}[nosep,leftmargin=1.4em,itemsep=1pt]
  \item \texttt{persuasiveness}: how convincingly does the argument make
    its case?
  \item \texttt{coherence}: is it logically structured with smooth
    transitions?
  \item \texttt{factual\_accuracy}: are claims accurate and
    well-supported by evidence?
  \item \texttt{value\_alignment}: does it invoke universal human values
    effectively?
  \item \texttt{rhetorical\_balance}: does it appropriately blend logos,
    pathos, and ethos?
  \item \texttt{stance\_alignment} (CRITICAL): does the argument actually
    argue in the stated direction? Score 5 if perfectly aligned, 0 if
    it argues the opposite. \textbf{If} $<$3, \texttt{overall} must
    be $\leq 2.0$.
  \item \texttt{overall}: holistic quality based on the above aspects (NOT a simple average)
\end{itemize}
\vspace{3pt}
Be critical but constructive. Identify specific strengths and weaknesses.

Output should be a JSON.
\end{tcolorbox}
\caption{Quality Evaluator system prompt. We only use the overall scores.
  Evaluation model is fixed to GPT-5.4.}
\label{fig:prompt-eval}
\end{figure*}


\begin{figure*}[h]
\begin{tcolorbox}[
  colback=gray!5, colframe=black!40, arc=3pt,
  title={\small\textbf{ELO Pairwise Judge --- System Prompt (GPT-5.4)}},
  fonttitle=\small, left=5pt, right=5pt, top=3pt, bottom=3pt
]
\small
You are an expert argument quality judge trained in rhetoric and
argumentation theory. Score Argument A and Argument B
\textit{independently} on each dimension (0--10):
\begin{itemize}[nosep,leftmargin=1.4em,itemsep=1pt]
  \item \texttt{persuasiveness}: how convincingly does it argue the
    position?
  \item \texttt{coherence}: how logically structured and fluent?
  \item \texttt{factual\_accuracy}: claims accurate and supported?
  \item \texttt{rhetorical\_balance}: appropriate combination of
    reasoning, emotional appeal, and credibility?
  \item \texttt{value\_alignment}: connects to audience values?
  \item \texttt{overall}: holistic quality based on the above aspects (NOT a simple average)
\end{itemize}
\vspace{3pt}
Respond with ONLY a JSON object with keys \texttt{"A"}, \texttt{"B"}
(each a dict of the six dimension scores), and \texttt{"reasoning"}
(one short passage: key differentiator between A and B). Do NOT inflate
scores --- use the full 0--10 range.\\[3pt]
Each pair is judged \textbf{twice} (orders A$\to$B and B$\to$A);
per-dimension scores are averaged across orderings before computing the
weighted aggregate and ELO outcome.
\end{tcolorbox}
\caption{ELO Pairwise Judge system prompt. Position bias is cancelled by
  averaging two score matrices (forward and reverse order). A score gap
  $>$\,0.5 is required to declare a win, preventing noise-driven ELO
  drift.
  We only use the overall score.
  }
\label{fig:prompt-elo}
\end{figure*}

\section{Evaluation Protocol Details}
\label{app:eval_details}

\subsection{Absolute Scoring}
Each argument is scored independently by GPT-5.4 using the Quality
Evaluator prompt (Figure~\ref{fig:prompt-eval}). 

For ELO pairwise ranking, we run pairwise comparisons per proposition. Each pair is
judged twice (A$\to$B and B$\to$A). Outcome: if
$w_A - w_B > 0.5$ then win; $<-0.5$ then loss; otherwise tie.

\subsection{Human Evaluation Details}
\label{app:human_eval}

We randomly sample 30 propositions and, for each, collect the arguments generated by ARGUS and the four baselines under a shared GPT-5-mini backbone. Each proposition is
independently ranked by 3 annotators. All annotators are college students majoring in computer and data science, are are proficient English speakers.

In particular, they are asked to rank the five arguments for a given proposition by
overall persuasive preference. The instructions directed them to weigh the following aspects:
\begin{itemize}
  \item \emph{Structure and coherence:} whether the argument is well organized
        and logically fluent.
  \item \emph{Persuasiveness:} whether the argument is convincing and more
        likely to move the reader and shift their opinion.
  \item \emph{Rhetorical technique:} whether the argument employs a variety of
        devices (e.g., concrete examples, quantitative evidence) to strengthen
        its overall persuasive effect.
\end{itemize}
For each proposition, annotators produced a full ranking of the five arguments and a short free-text rationale explaining their ordering.

We measure agreement using Kendall's $\tau_b$ as annotators may assign ties. For each proposition we compute
$\tau_b$ between every pair of annotators' rankings and average across pairs; the overall agreement score is 0.242. Given the subjectivity of evaluating persuasion on controversial propositions, this reflects a positive level of agreement.
\end{appendix}

\end{document}